\documentclass[twoside,11pt]{article}

\usepackage{jmlr2e}
\usepackage[utf8]{inputenc}

\usepackage{bm}

\usepackage{amsmath}
\usepackage{dsfont}
\usepackage{mathptmx}
\usepackage{amsfonts}
\usepackage{subfig}
\usepackage{tabularx}
\usepackage{multirow}
\usepackage{url}
\usepackage{color}

\jmlrheading{-}{-}{-}{-}{-}{-}

\ShortHeadings{Bayesian Nonparametric Crowdsourcing}{Moreno,  Teh, Perez-Cruz and Artés-Rodríguez}
\firstpageno{1}

\begin{document}

\title{Bayesian Nonparametric Crowdsourcing}

\author{\name Pablo G. Moreno \email pgmoreno@tsc.uc3m.es \\
       \addr Department of Signal Theory and Communications\\
      University Carlos III of Madrid\\
       Avda. de la Universidad, 30\\
       28911 Leganés (Madrid, Spain)
       \AND
       \name Yee Whye Teh \email y.w.teh@stats.ox.ac.uk \\
       \addr Department of Statistics\\
       1 South Parks Road\\
       Oxford OX1 3TG, UK
       \AND
       \name Fernando Perez-Cruz \email fernando@tsc.uc3m.es \\
       \addr Department of Signal Theory and Communications\\
      University Carlos III of Madrid\\
       Avda. de la Universidad, 30\\
       28911 Leganés (Madrid, Spain)
       \AND
       \name Antonio Artés-Rodríguez \email antonio@tsc.uc3m.es \\
       \addr Department of Signal Theory and Communications\\
      University Carlos III of Madrid\\
       Avda. de la Universidad, 30\\
       28911 Leganés (Madrid, Spain)}

\editor{-}

\maketitle

\begin{abstract}
Crowdsourcing has been proven to be an effective and efficient tool to annotate large datasets. User annotations are often noisy, so methods to combine the annotations to produce reliable estimates of the ground truth are necessary.
We claim that considering the existence of clusters of users in this combination step can improve the performance. This is especially important in early stages of crowdsourcing implementations, where the number of annotations is low. At this stage there is not enough information to accurately estimate the bias introduced by each annotator separately, so we have to resort to models that consider the statistical links among them.  In addition, finding these clusters is interesting in itself as knowing the behavior of the pool of annotators allows implementing efficient active learning strategies. Based on this, we propose in this paper two new fully unsupervised models based on a Chinese Restaurant Process (CRP) prior and a hierarchical structure that allows inferring these groups jointly with the ground truth and the properties of the users. Efficient inference algorithms based on Gibbs sampling with auxiliary variables are proposed. Finally, we perform experiments, both on synthetic and real databases, to show the advantages of our models over state-of-the-art algorithms.

\end{abstract}


\section{Introduction}
\label{Introduction}

Crowdsourcing services are becoming very popular as a mean of outsourcing tasks to a large crowd of users. The best-known tool is Mechanical Turk \cite[]{MechanicalTurk}, in which \textit{Requesters} are able to post small tasks for \textit{Providers} registered in the system, who complete them for a monetary payment set by the Requester. In machine learning, crowdsourcing allows to distribute the labeling of a dataset among a pool of users, so each user only labels a subset of the instances. The advantage is the ability to  gather large datasets in a short time and, generally, at a low cost. Successful examples include, but it is not limited to, LabelMe \cite[]{LabelMe:2008} or GalazyZoo \cite[]{Galaxy:2010}. 

The quality of the labels retrieved by crowdsourcing is uneven. Unlike traditional ways of gathering a labeled dataset in which labels are provided by a small set of motivated experts, we deal now with a large number of users who are not necessarily experts nor motivated. Further, we might have little or no information about them to perform quality control tests. This motivates the development of statistical models for reliably estimating ground truth from noisy and biased labels provided by users.

Another problem that has received significant attention of late, is the detection of groupings among the labelers \cite[]{Simpson:11,Simpson:2013}. In most crowdsourcing applications we can identify several types of users: experts, novices, spammers and even malicious or adversarial annotators. Identifying these groups of users and learning about their properties is useful to design efficient crowdsourcing strategies that minimize the overall cost, selecting the most suitable users for a labeling task. For example, if we could identify spammers we could ban them from the system and avoid wasting resources. In the same way, if a user is identified as an expert in a particular task, we could reward him by increasing his pay-off or giving him preference over other users when the time to select new tasks comes. 

Usually, the detection of grouping of labelers is tackled in a post processing step, after the ground truth has been estimated from user annotations (see Section \ref{RelatedWork}). In particular \cite[]{Simpson:11} were the first to tackle the join problem. They estimated the  ground truth using a previous model called iBBC by \cite[]{Kim:12} and then, as a post-processing step they infer the different clusters of users. Therefore, the estimation of the ground truth is done without considering the clustering structure of the users. 

In this paper, we propose two unsupervised Bayesian nonparametric models to combine the labels provided by the users in a crowdsourcing scenario, taking into account the presence of clusters of users. Our models jointly solve the problem of the estimation of ground truth and the problem of identification of clusters of users and their properties.  The estimation of the ground truth improves the clustering of the users and vice-versa, thus performing better than current state-of-the-art, namely \cite[]{Kim:12,Simpson:11}. The overall improvement in both tasks is particularly important in the early stages of a crowdsourcing project, when the number of annotations provided by the users is very low. In this case, algorithms that estimate the properties of each user independently, without considering the dependencies among them, tend to provide poor estimates, and may perform worse than majority voting (see Section \ref{Results}).  

In the first model, we propose a clustering structure using a CRP prior \cite[]{Pitman02} which allows flexible modeling of the number of clusters of users. In this model, all the users that belong to the same cluster share the same parameters governing the way they label instances, and therefore, they have the same behavior. Forcing all the users to share the same exact parameters, is a strong assumption that might lead to groupings with a large number of cluster. Therefore, these groupings are difficult to interpret and not very useful. To relax this assumption we propose a second model in which users that belong to the same cluster are modeled as having similar parameters, but allows each user to have its own parameters using a hierarchical Bayesian approach.

In this paper, we rely on a Bayesian nonparametric model, because we are not only interesting in having an accurate model but also in having an interpretable one. In the experiments (Section \ref{Results}) we show that the error rates between the two proposed models are not significantly different. However, the second one is interpretable, in the sense that it perfectly identifies each kind of clusters and it reports the least number of them. The interpretability of the model is principal to us, because we want to use the model to identify the ‘good’ annotators and be able to reward them accordingly, while other models are not able to provide this information. 

The rest of the paper is organized as follows. In Section \ref{Models}, we present the  two new generative models for crowdsourcing that take into account the clustering structure of the users. In Section \ref{Inference}, we propose efficient Markov chain Monte Carlo (MCMC) inference algorithms for estimating the different groups of users as well as the ground truth.  In Section \ref{RelatedWork}, we review related literature on crowdsourcing and the identification of user clusters in the context of crowdsourcing. In Section \ref{Results},  we validate our model on synthetic data and we perform several experiments on real datasets to show the advantages of our models over state-of-the-art algorithms. Finally, we conclude this paper in Section \ref{Conclusions} and present possible extensions for the future.

\section{Hierarchical Bayesian Combination of Classifiers}
\label{Models}

In this section, we propose two different models. Both algorithms receive as input a set of noisy labels $\bm{Y} \in \{0,...,C\}^{N \times L}$ provided by $L$ users for $N$ instances. The element $y_{i\ell}$ represents the label given by the user $\ell$ to the instance $i$ and it is $0$ if the user did not label the corresponding instance. Notice that this matrix $\bm{Y}$ is highly sparse in the early stages of a crowdsourcing application. This is known as the \textit{cold start problem} \cite[]{Schein:2002}, i.e. the difficulty of drawing any inferences due to the lack of information. Notice that $\bm{Y}$ is the only observed variable in the models.

The output of the algorithms is the set of true but unknown labels of the instances $\bm{z} \in \{1,...,C\}^N$, where $z_{i}$ indicates the true label estimate of  the instance $i$. 

We denote by $\left[ L \right]=\{1,2,...,L\}$ the set of indices of the users and by $\bm{\pi}_L$ a partition of $\left[ L \right]$. A partition is a collection of mutually exclusive, mutually exhaustive and non-empty subsets called clusters. We denote the cluster assignment of the user $\ell$ with a variable $q_\ell$ such that $q_\ell = m$ denotes the event that the user $\ell$ is assigned to cluster $m \in \bm{\pi}_L$ . 

\subsection{Clustering based Bayesian Combination of Classifiers}

\begin{figure}
\centering
\begin{tabular}{ccc}
\subfloat[iBCC model]{{\label{fig-baseline_model}}\includegraphics[width=0.44\textwidth]{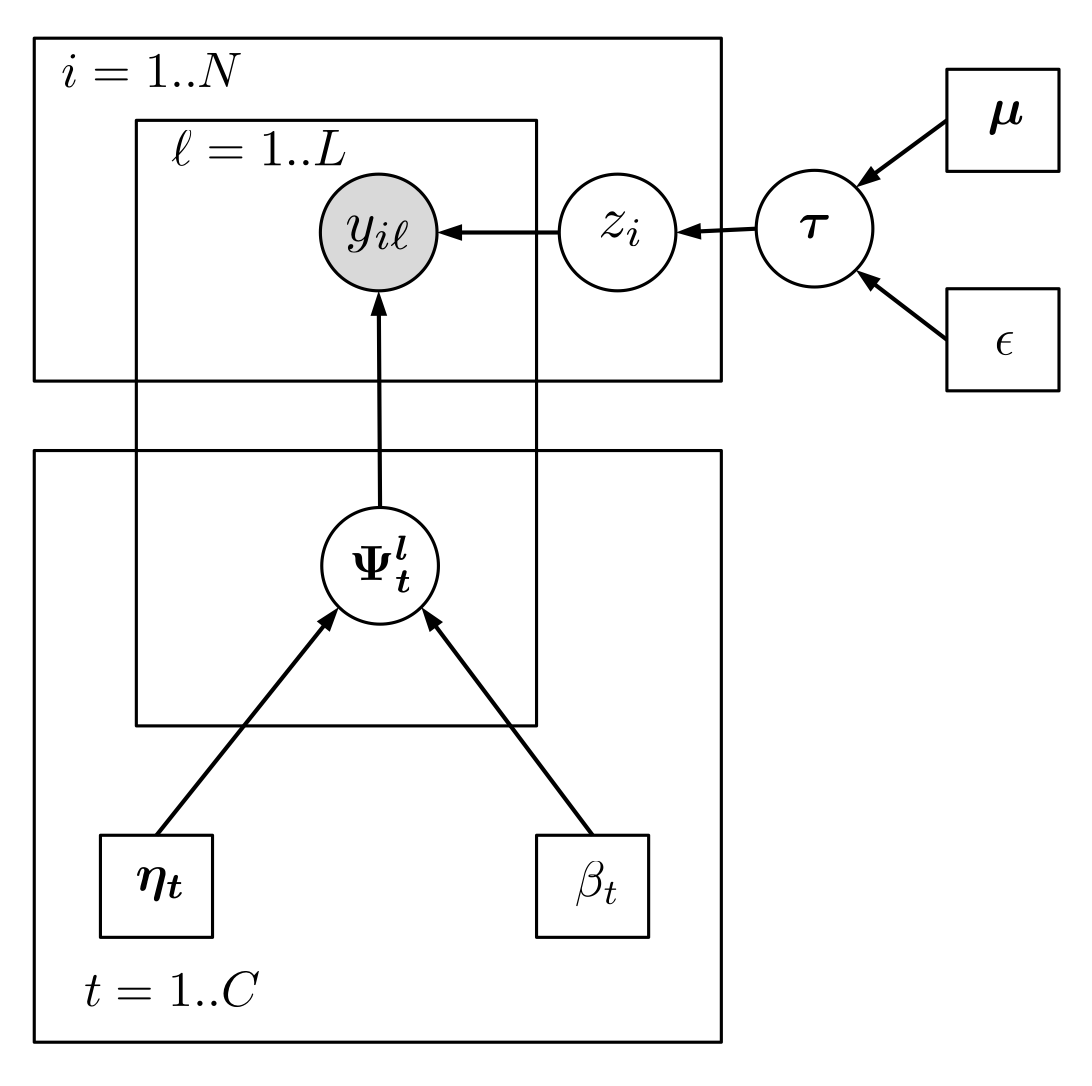}} \hspace{5mm}
\subfloat[cBCC model]{{\label{fig-first_model}}\includegraphics[width=0.45\textwidth]{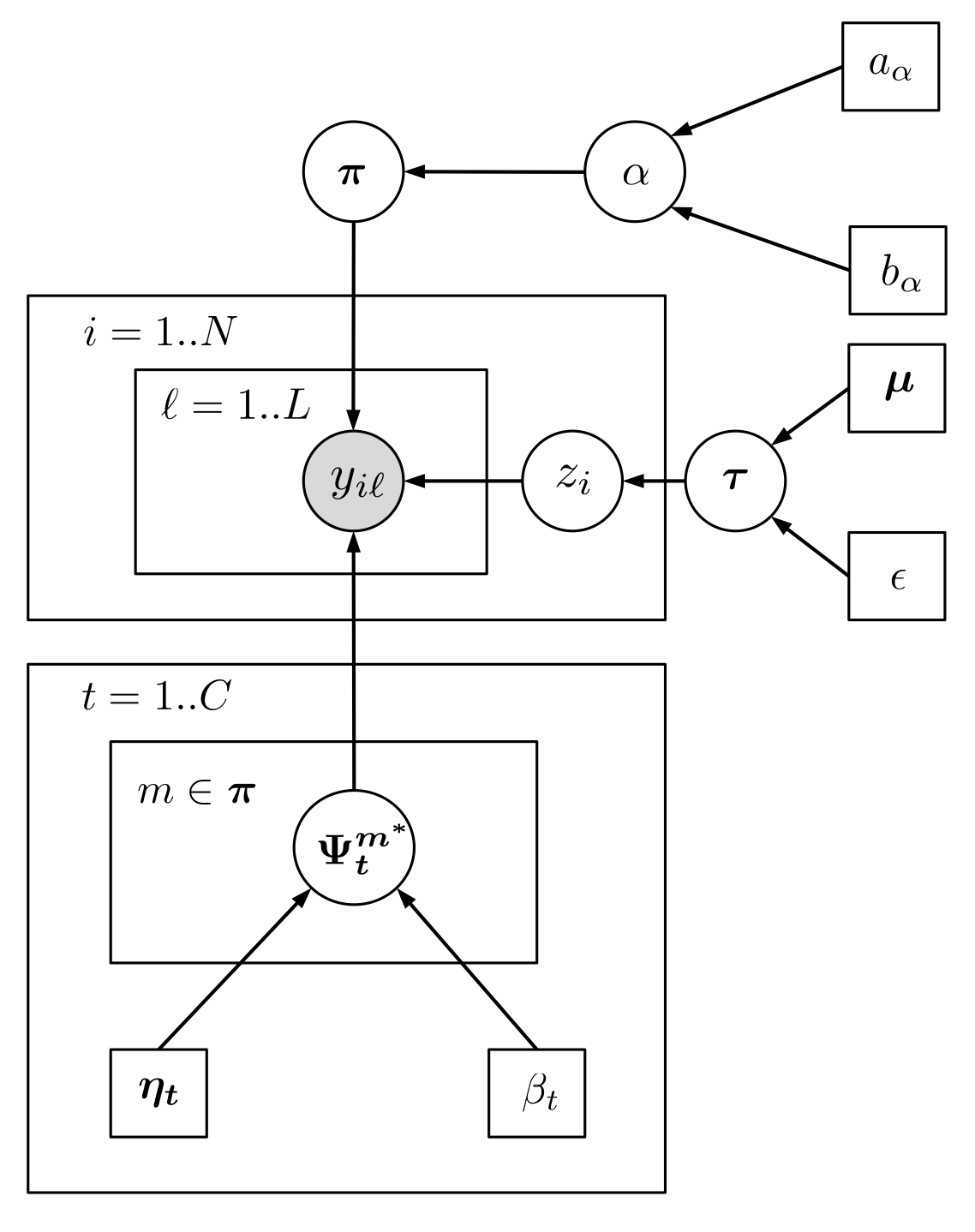}} 
\end{tabular}
\caption{Graphical model representation of the iBCC and cBCC models}
\end{figure}

Firstly, we propose a model for users in which they can belong to different clusters. In each cluster all the users have the same properties. We name it Clustering based Bayesian Combination of Classifiers (cBCC) (see Figure \ref{fig-first_model}) and it has the following observation model:
\begin{align*}
&y_{i\ell} | z_i,\bm{\pi}, \bm{\Psi} \stackrel{i.i.d}{\sim} \mathrm{Discrete}(\bm{\Psi}_{z_i}^{q_\ell})\\ 
&z_i | \bm{\tau} \stackrel{i.i.d}{\sim} \mathrm{Discrete}(\bm{\tau}).
\end{align*}
We assume that all the users that belong to cluster $m \in \bm{\pi}$ share the same properties, i.e. the same confusion matrix $\bm{\Psi}^m \in [0,1]^{C \times C}$, where $\Psi_{tc}^m$ is the probability that a user allocated in cluster $m$ labels an instance as $y = c$  when the ground truth is $z = t$. We use the notation $\bm{\Psi}_{t}^m \in \mathcal{S}^C$ to denote the row $t$ of $\bm{\Psi}^m$, where $\mathcal{S}^C$ is the C-dimensional probability simplex. The component $\tau_t$ of  $\bm{\tau} \in \mathcal{S}^{C}$ is the probability of the ground truth $z$ being equal to the category $t \in \{1,...,C\}$. 

We also need to define priors to complete the Bayesian model. In particular, we choose conjugate priors:
\begin{align*}
&\bm{\Psi}_t^m| \bm{\beta}, \bm{\eta} \sim \mathrm{Dir}(\beta_t \bm{\eta_t}) \\
&\bm{\tau}| \epsilon, \bm{\mu} \sim\mathrm{Dir}(\epsilon \bm{\mu}),
\end{align*}
where we use a Dirichlet prior on each of the rows of the confusion matrices in which $\bm{\eta}_t \in \mathcal{S}^{C}$ is the mean value of $\bm{\Psi}_{t}^m$ while $\beta_t \in \mathbb{R}_+$ is related to its precision. Notice that this is an over parametrization of the Dirichlet distribution, which only needs $C$ parameters to be fully determined. However, this decomposition is useful to interpret the results as well as for the development of the inference algorithms in Section \ref{Inference}. Likewise,  we set a Dirichlet prior on $\bm{\tau}$, where $\bm{\mu}\in \mathcal{S}^{C}$ is the mean and $\epsilon\in \mathbb{R}_+$ relates to the precision. 

We could use a parametric model in which the cardinality of the partition $M = |\bm{\pi}|$ is fixed a priori. Unfortunately, in this case the inferences are sensitive to the value of $M$ chosen. In the limiting case $M=1$ the model is equivalent to majority voting. If M is too large the model does not take advantage of the presence of clusters of users. In the limiting case $M=L$ each user becomes a singleton cluster, and the model does not capture the dependencies among the users.

To find $M$ we could use  traditional model selection strategies like cross-validation \cite[]{Stone:1974} or Bayesian Information Criterion \cite[]{BIC98}. This approach has two limitations. First, we usually do not have access to a validation set for which $\bm{z}$ is known. Second, is the high computational complexity.  An alternative pathway is to set a prior on the space of partitions. We denote by $\mathcal{P}_L$ the space of all partitions $\pi_L \in \mathcal{P}_L$. In a Bayesian setting, we have to set the prior without observing the number of users. One option is to set a prior on an infinite number of users, i.e. on partition $\bm{\pi} \in \mathcal{P}_\infty$. To build such a prior we further assume that the observations are exchangeable and that we deal with consistent random partitions (see \cite[]{Pitman02}). A (exchangeable and consistent) prior on $\mathcal{P}_\infty$  is the CRP introduced in \cite[]{blackwell:1973} and that can be seen as the induced distribution over the partition space by a Dirichlet Process (DP) \cite[]{Ferguson:1973, Antoniak:1974}. We place a CRP prior over the users' partions:  

\begin{equation}
\label{eq:jointCRP}
\bm{\pi} | \alpha,d  \sim \mathrm{CRP}(\alpha).
\end{equation}

We can generate samples from this prior using the following conditional distributions:
\begin{equation}
p(q_\ell = m | \bm{\pi}^{\neg \ell}, \alpha) \propto
\begin{cases}
|m|^{\neg \ell} , \hspace{5mm} &m \in \bm{\pi}^{\neg \ell}\\
\alpha , \hspace{5mm}&m = \emptyset,
\end{cases}
\end{equation}
where $|m|$ represents the number of users in cluster $m$ and $|m|^{\neg \ell}$ is equal to $|m|$ excluding user $\ell$. We denote by $\bm{\pi}^{\neg \ell}$ the partition with the user $\ell$ removed and $q_\ell = \emptyset$ denotes the event that user $\ell$ is assigned to a new cluster.  $\alpha$ is the so called concentration parameter and control the a priori probability of generating new clusters. We further place a gamma prior over the concentration parameter $\alpha$:

\begin{equation}
\label{eq:alphad}\alpha | a_\alpha, b_\alpha \sim \mathrm{Gamma}( a_\alpha, b_\alpha).
\end{equation}

If  $\alpha$ tends to infinity, every user is allocated to a singleton cluster. If $\alpha$ tends to 0, all the users share the same confusion matrix and the model produces a majority voting solution

In general, the CRP assigns more mass to partitions with a small number of clusters. This is sensible in our case, because when the number of annotations is scarce, majority voting may perform better than more elaborate algorithms since there is no enough information to estimate the individual properties of the users. In this case, the CRP prior dominates and therefore all users are allocated to the same cluster. When the number of annotations increase, the likelihood term dominates and different clusters of users are created.


Finally, we analyze the correlation structure that it is introduced among the users as a consequence of this clustering.  The correlation a priori among two users $\ell$ and $\ell'$ is:
\begin{equation}
\label{eq:correlation}
\begin{split}
&\mathrm{Corr}(\mathds{I}(y_{i\ell}=a),\mathds{I}(y_{i\ell'}=b) | z_i = t)  =
\begin{cases}
 - \left( \frac{1}{1+\alpha} \right) \left(   \frac{1}{1 + \beta_t} \right) \sqrt{ \frac{ \eta_{ta} \eta_{tb} }{  \left(  1 - \eta_{ta} \right) \left(  1 - \eta_{tb} \right) } } \hspace{5mm} &a \neq b\\
 \left( \frac{1}{1+\alpha} \right) \left(   \frac{1}{1 + \beta_t} \right)  \hspace{5mm} &a = b.
 \end{cases}
 \end{split}
\end{equation}
Here, $\mathds{I}(.)$ represents the indicator function. The proof is in the supplementary material. In Section \ref{RelatedWork}, we show how this model relates to other state-of-the-art algorithms. 

\subsection{Hierarchical Clustering based Bayesian Combination of Classifiers}

In the cBCC model, the users that belong to the same cluster share the same confusion matrix. However, in a practical situation, each user has a behavior that is different from  every other user, but it is in some sense similar to the behavior of users that are allocated to its cluster.

\begin{figure}
\centering
\includegraphics[width=0.46\textwidth]{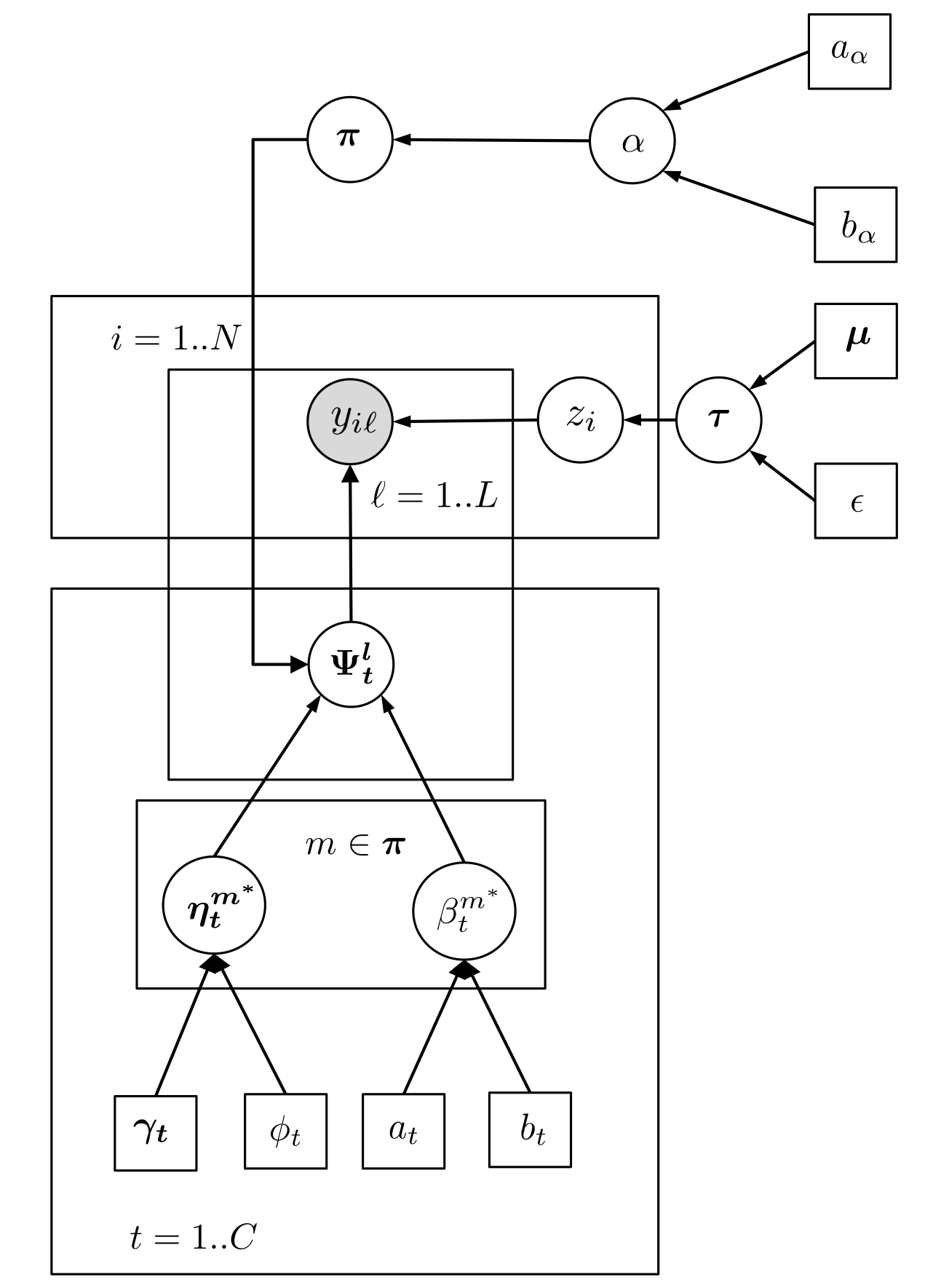}
\caption{Graphical model representation of the hcBCC model}
\label{fig:second_model1}
\end{figure}

To capture this behavior, we propose a hierarchical extension of the cBCC model called hcBCC (hierarchical cBCC) depicted in Figure \ref{fig:second_model1}. The observation model is the following:

\begin{align*}
&y_{i\ell} | z_i, \bm{\Psi} \stackrel{i.i.d}{\sim} \mathrm{Discrete}(\bm{\Psi}_{z_i}^{\ell}) \\
&z_i | \bm{\tau} \stackrel{i.i.d}{\sim} \mathrm{Discrete}(\bm{\tau}).
\end{align*}

Now each user has its own confusion matrix $\bm{\Psi}^{\ell}$ in contrast to the cBCC model where we had a confusion matrix per cluster $\bm{\Psi}^{m}$. To capture the similarity between users that belong to the same cluster we use the following hierarchical prior:

\begin{align*}
&\bm{\Psi}_{t}^\ell | \bm{\pi},\bm{\beta}, \bm{\eta} \sim \mathrm{Dir}(\beta_t^{q_\ell} \bm{\eta}_{t}^{q_\ell}) \\
&\beta_t^m| a_t, b_t \sim \mathrm{Gamma}(a_t, b_t)  \\
&\bm{\eta}_{t}^m| \bm{\phi}, \bm{\gamma} \sim \mathrm{Dir}(\phi_t \bm{\gamma}_{t})\\
&\bm{\tau} | \epsilon, \bm{\mu} \sim \mathrm{Dir}(\epsilon \bm{\mu}).
\end{align*}

In this way, the confusion matrices of all users that belong to the same cluster $m$  are generated from the same distribution. In particular, each of the rows of the confusion matrices of all the users that belong to cluster $m$, i.e. $\lbrace \bm{\Psi}_{t}^\ell : q_{\ell} = m \rbrace$, are i.i.d. samples from the same Dirichlet distribution whose parameters are $\beta_t^m$ and $\bm{\eta}_{t}^m$.  A Dirichlet prior is set on the vector $\bm{\eta}_{t}^m$ while a gamma prior is set on the scalar $\beta_t^m$. Finally, for $\bm{\pi}$ and $\alpha$ we, respectively, use the same priors given by Equations \ref{eq:jointCRP} and \ref{eq:alphad}. With this we have a model where we no longer cluster the confusion matrices of the users, but the distributions that generate them. 

Notice that the vector $\bm{\beta}^m$ governs the variability among the users that belong to the same cluster $m$. The bigger are these values, the lower is the intra-cluster variability. If we make each of the components of $\bm{\beta^m}$ tend to infinity, then the variability among the users tend to $0$ and the model becomes equivalent to the cBCC model. In this way, this model can be seen as a generalization of some state-of-the-art methods (see Section \ref{RelatedWork}).

\section{Inference}
\label{Inference}

Computing the posterior distribution of the clusters allocation, the properties of the users and the estimated ground is intractable, so we have to resort to approximate inference.  Since the proposal of the DP by \cite[]{Ferguson:1973}, approximate inference schemes based on Markov Chain Monte Carlo (MCMC) methods have played a crucial role: \cite[]{Escobar:1994,MacEachern1998,Neal:2000,Ishwaran2001,walker07,KGW11} among others. In this section we propose to use Gibbs sampling together with the corresponding auxiliary variables whenever it is not possible to compute the conditional distributions due to non-conjugacies.

\subsection{cBBC} 
\label{cBBC_inference} 

We use a collapsed Gibbs sampling algorithm where we integrate out the variables $\bm{\Psi}^m$ and $\bm{\tau}$, obtaining the following new set of equations:
\begin{align*}
&p(\bm{Y} | \bm{\pi}, \bm{z}, \bm{\eta}, \bm{\beta})= \prod_m \prod_t \left[ \frac{\Gamma(\beta_t)}{\Gamma(n_{mt} + \beta_t)} \prod_c \frac{\Gamma(n_{mtc} + \beta_t \eta_{tc})}{\Gamma(\beta_t \eta_{tc})} \right],\\ &
p(\bm{z}| \epsilon, \bm{\mu})= \frac{\Gamma(\epsilon)}{\Gamma(N + \epsilon)} \prod_t \frac{\Gamma(n_m + \epsilon \mu_t)}{\Gamma(\epsilon \mu_t)},
\end{align*}
where $\Gamma(\cdot)$ denotes the gamma function.  We denote $n_{i\ell mtc}=\mathds{I}(z_i=t, y_{i\ell}=c, y_{i\ell} \neq 0 , q_\ell=m)$, and when an index of this variable is omitted we assume it is summed out. For example, $n_{mtc}$ represents the number of annotations equal to $c$ provided by the users of cluster $m$ for the set of instances that belong to category $t$. We use Gibbs sampling to infer the value of the ground truth $\bm{z}$, the clusters of annotators $\bm{\pi}$, as well as the hyper parameters of the CRP, conditioned on the observed variables $\bm{Y}$. 

Firstly, to update the cluster assignment of annotator $\ell$, we need the conditional distribution of $q_\ell$ given the rest of the variables:
\begin{align*}
&p(q_\ell = m | 
\text{rest}
) \propto
\begin{cases}
 \displaystyle n_m^{\neg \ell} \times  \prod_t \frac{\Gamma(n_{mt}^{\neg \ell} + \beta_t)}{\Gamma(n_{mt}^{\neg \ell} + n_{\ell t}  + \beta_t)}  \prod_t \prod_c \frac{\Gamma(n_{mtc}^{\neg \ell} + n_{\ell tc}  + \beta_t \eta_{tc})}{\Gamma(n_{mtc}^{\neg \ell} + \beta_t \eta_{tc})}, \hspace{7mm} &m \in \bm{\pi}^{\neg \ell}\\
\displaystyle \alpha \times \prod_t \frac{\Gamma(\beta_t)}{\Gamma(n_{\ell t} + \beta_t)}  \prod_t \prod_c \frac{\Gamma(n_{\ell tc} + \beta_t \eta_{tc})}{\Gamma(\beta_t \eta_{tc})}, \hspace{7mm} & m = \emptyset
\end{cases}
\end{align*}
where $q_\ell = \emptyset$ denotes the event that user $\ell$ is assigned to a new cluster. The quantities $n_{mt}^{\neg \ell}$ and $n_{mtc}^{\neg \ell}$ are defined in the same way as $n_{mt}$ and $n_{mtc}$ respectively, but excluding the annotator $\ell$.

To sample the estimate of the ground truth $z_i$ of each instance conditioned on the rest of the variables, the required conditional distribution is:
\begin{equation}
\begin{split}
&p(z_i = t | \text{rest} 
) \propto \left( n_t^{\neg i} + \epsilon \mu_t \right)  \times 
\prod_m \left[ \frac{\Gamma(n_{mt}^{\neg i} + \beta_t )}{\Gamma(n_{mt}^{\neg i} + n_{im} +  \beta_t )} \prod_c \frac{\Gamma(n_{mtc}^{\neg i} + n_{imc} +  \beta_t \eta_{tc} )}{\Gamma(n_{mtc}^{\neg i} + \beta_t \eta_{tc})} \right]
\end{split}
\end{equation}
The quantities $n_{mt}^{\neg i}$ and $n_{mtc}^{\neg i}$ again correspond to $n_{mt}$ and $n_{mtc}$ but excluding the instance $i$. Finally, we sample the concentration parameter $\alpha$ following the procedure in \cite[]{Escobar:1994}.


\subsection{hcBCC}
\label{hcBBC_inference} 

As in the cBCC we start by  integrating out the $\bm{\Psi}^\ell$ and $\bm{\tau}$ variables:
\begin{align*}
&p(\bm{Y} | \bm{z}, \bm{\pi} , \bm{\eta}, \bm{\beta}) =\prod_{m} \prod_{\ell: q_\ell=m} \prod_{t}  \left[ \frac{\Gamma(\beta_t^m)}{\Gamma(n_{\ell t} + \beta_t^m)} \prod_c \frac{\Gamma(n_{\ell tc} + \beta_t^m \eta_{tc}^m)}{\Gamma(\beta_t \eta_{tc}^m)}\right], \\
&p(\bm{z}| \epsilon, \bm{\mu})= \frac{\Gamma(\epsilon)}{\Gamma(N + \epsilon)} \prod_t \frac{\Gamma(n_m + \epsilon \mu_t)}{\Gamma(\epsilon \mu_t)}
\end{align*}
The variables we need to sample from are $\bm{\pi}$ and the ground truth estimate $\bm{z}$.  Note however that we cannot marginalize out the cluster parameters $\bm{\eta}$ and $\bm{\beta}$, as the Dirichlet prior and the Gamma prior are not conjugate to the likelihoods given above, so that these variables will have to be sampled as well.

The conditional distribution of $p(q_\ell=m|rest)$ when $m \in \bm{\pi}^{\neg \ell}$ can be computed like in the cBCC model. However, to compute $p(q_\ell=m|rest)$ when $m = \emptyset$ we need to integrate the parameters of the new clusters, i .e . $\bm{\beta}$ and $\bm{\eta}$. In this case, due to the non-conjungancy we cannot solve this integral analytically. Instead, we use the recently proposed \textit{Reuse algorithm} by \cite[]{favaro:2012}. This algorithm is similar to the well-known Algorithm 8 in \cite[]{Neal:2000}, where the idea is to use a set of $h$ auxiliary empty clusters $H_{empty}$ to approximate the integral. However, the \textit{reuse algorithm} is more efficient as it requires less simulations from the prior over the cluster parameters.  For each cluster $m\in \bm{\pi}\cup H_{empty}$ we keep track of the parameters $\bm{\beta}^m$ and $\bm{\eta}^m$.  The conditional distribution of $q_\ell$ is then:
\begin{equation}
\begin{split}
\label{eq:algorithm8}
&p(q_\ell = m | \text{rest}
) \propto
\begin{cases}
n_m^{\neg \ell} \times  \prod_t \frac{\Gamma(\beta_t^m)}{\Gamma(n_{\ell t}  + \beta_t^m)} \prod_t \prod_c \frac{\Gamma( n_{\ell tc}  + \beta_t^m \eta_{tc}^m)}{\Gamma(\beta_t^m \eta_{tc}^m)}, \hspace{7mm} &m \in \bm{\pi}^{\neg \ell}\\
\frac{\alpha}{h} \times \prod_t \frac{\Gamma(\beta_t^m)}{\Gamma(n_{\ell t}  + \beta_t^m)} \prod_t \prod_c \frac{\Gamma( n_{\ell tc}  + \beta_t^m \eta_{tc}^m)}{\Gamma(\beta_t^m \eta_{tc}^m)}, \hspace{7mm} &m \in H_{empty}
\end{cases}
\end{split}
\end{equation}
If an auxiliary empty cluster is chosen, it is moved into the partition $\bm{\pi}$, and a new empty cluster is created in its place by sampling from the prior over cluster parameters.  If a cluster in $\bm{\pi}$ is emptied as a result of sampling $q_\ell$, it is moved into $H_{\empty}$, displacing one of the empty clusters (picked uniformly at random).  In addition, at regular intervals the parameters of the empty clusters are refreshed by simulating them from their priors, while those in $\bm{\pi}$ are updated.

Again, due to the non-conjugacy of the Dirichlet and Gamma priors, the conditional distributions of the parameters $\bm{\eta}^m$ and $\bm{\beta}^m$ for $m\in\bm{\pi}$ cannot be computed analytically. 
To solve this, we use an auxiliary variable method similar to \cite[]{Escobar:1994} and \cite[]{Teh:2003}. Specifically, we introduce two auxiliary variables $\bm{\nu}$ and $\bm{s}$ (see the supplementary material for further details), and apply the following Gibbs updates that leave invariant the posterior distribution:
\begin{align*}
& \nu_{\ell t} \sim \mathrm{Beta}(\beta_t^{q_\ell}), \hspace{7mm} s_{\ell t c} \sim \mathrm{Antoniak}(n_{\ell t c}, \beta_t^{q_\ell} \eta_{tc}^{q_\ell})\\
&\eta_{t:}^m \sim \mathrm{Dir}\left(\sum_{\{ \ell:q_\ell = m \}} s_{\ell t:} + \phi_t \gamma_{t:}\right), \\
&\beta_t^m \sim \mathrm{Gamma}\left(\sum_{\{ \ell:q_\ell = m \}} \sum_c s_{\ell tc} + a_t, b_t - \sum_{\{ \ell:q_\ell = m \}} \log(\nu_{\ell t}) \right)
\end{align*}
Here the Antoniak distribution introduced in \cite[]{Antoniak:1974} is simply the distribution of the number of clusters in a partition  of $n_{\ell t c}$ items under a CRP with concentration parameter $\beta_t^{q_\ell} \eta_{tc}^{q_\ell}$. 


To update $z_i$, we compute its conditional distribution given the rest of the variables:
\begin{equation}
\begin{split}
&p(z_i = t | \text{rest}
) \propto \left( n_t^{\neg i} + \epsilon \mu_t \right)  
 \prod_m \prod_{\{\ell: q_\ell=m\}} \frac{\prod_c  ( n_{\ell tc}^{\neg i} + \beta_t \eta_{tc})^{\mathds{I}(y_{i \ell}=c)}}{( n_{\ell t}^{\neg i} + \beta_t)^{\mathds{I}(y_{i \ell} \neq 0)}}\\
\end{split}
\end{equation}
Finally, we use the same scheme as the one applied in Section \ref{cBBC_inference} to update $\alpha$.

\section{Related Work}
\label{RelatedWork}

Dawid and Skene in a seminal work proposed a model in which each user is characterized by a confusion matrix, and they use the EM algorithm to estimate the most likely values of both the parameters governing the behavior of each user and the ground truth \cite[]{Dawid:79}.  Similar models have been applied to depression diagnosis \cite[]{Young83} and myocardial infarction \cite[]{Rindskopf86}, among others.

\cite[]{zoubin:03,Kim:12} propose a Bayesian extension of \cite[]{Dawid:79} called Independent Bayesian Combination of Classifiers (iBCC), whose graphical model is shown in Figure \ref{fig-baseline_model}. In our cBCC model, if $\alpha$ tends to infinity, every user is allocated in a different cluster, and it becomes equivalent to the iBBC model. We see that in this case, the correlation a priori among two users (Equation \ref{eq:correlation}) is zero. Also, in the hcBCC model, if each component of $\bm{\phi}$ tends to infinity, and we also make the quantities $a_t$ and $b_t$ tend to infinity with a fixed $\frac{a_t}{b_t}$ ratio for all $t$, then we recover the iBCC model with $\bm{\eta}_t = \bm{\gamma}_t$ and $\beta_t = \frac{a_t}{b_t}$. If $\alpha$ tends to $\infty$, then the model is equivalent to the iBCC model, but with additional priors on $\bm{\eta}$ and $\bm{\beta}$. To sum up, we can see each the cBCC and the iBCC as particularizations of the hcBCC model, which capture more complex relationships among the users. 

In \cite[]{zoubin:03,Kim:12} they also presented two extensions. The first one uses a latent variable that categorizes the instances in two classes: easy and difficult to classify. The assumption is that the annotators have the same behavior regarding the easy instances, while they are different for the difficult ones. In the second one, they propose a more flexible correlation model based on a factor graph. However, these models do not identify groupings of users.

In \cite[]{Simpson:11}  the authors extend \cite[]{zoubin:03,Kim:12} in two directions. First, they derived a variational inference algorithm for the iBCC, which is more efficient for large datasets. Second, they apply community detection algorithms to the estimated confusion matrices to detect clusters of users with the same behavior. Recently, they have extended the model to the case in which the properties of the users can vary in time \cite[]{Simpson:2013}.  In both cases, the detection of groups of users is made in a post-hoc manner and therefore, this information is not used to improve the estimation of the confusion matrices of the users or the ground truth estimate. To the extent of our knowledge, only \cite[]{Kajino13} performs the inference of the groups of users and the ground truth at the same time, using convex optimization. However, the performance depends on a constant that controls the strength of the clustering and for tuning this constant, the authors rely on a labeled validation set. Our algorithm, on the other hand, is fully unsupervised and therefore can be apply to the standard problem presented in \cite[]{zoubin:03,Kim:12}. 

Another research line that is related to the problem is  relaxing the assumption of the existence of one single gold standard, which is a limiting assumption when the tasks involved in the crowdsourcing problem are subjective and accept multiple reasonable answers \cite[]{Wauthier12,Tian13}. In this paper, we focus on a crowdsourcing scenario with a well defined gold standard that we aim to predict.

\section{Experiments}
\label{Results}

In this section, we firstly use synthetic datasets based on different assumptions to validate our models. In the second part we use publicly available real datasets to compare our two models with state-of-the-art algorithms highlighting their advantages.

\subsection{Synthetic datasets}
\label{synthetic}

We generate three different datasets following respectively the assumptions of the hcBCC, cBCC and iBCC models.  In order to analyze the properties of the algorithms, we apply each of them to each of the generated datasets.

\begin{figure}
\centering
\subfloat[Users' properties for dataset1]{%
\begin{minipage}[b][6.8cm][t]{.45\textwidth}
\centering
\vspace{4mm} 
\label{fig-DB_mean}\includegraphics[width=0.9\textwidth]{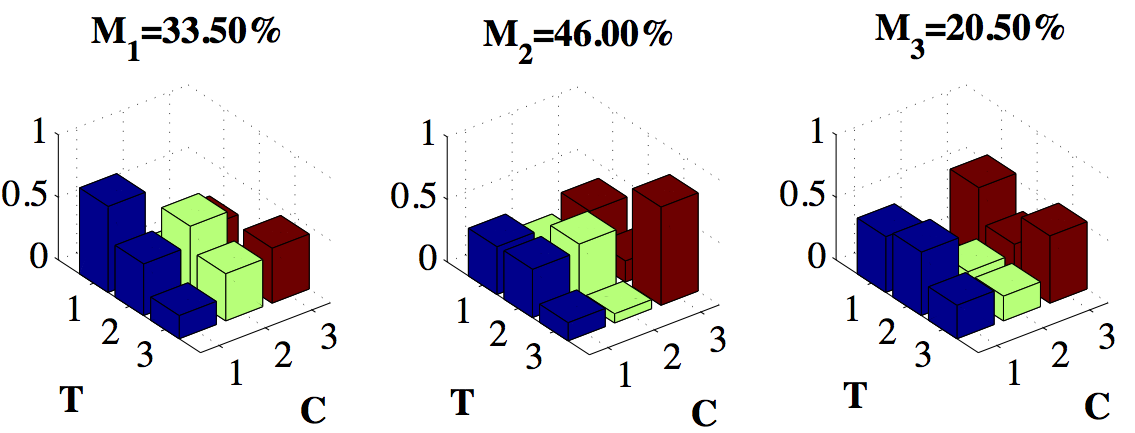}

\label{fig-DB_variance}\includegraphics[width=0.9\textwidth]{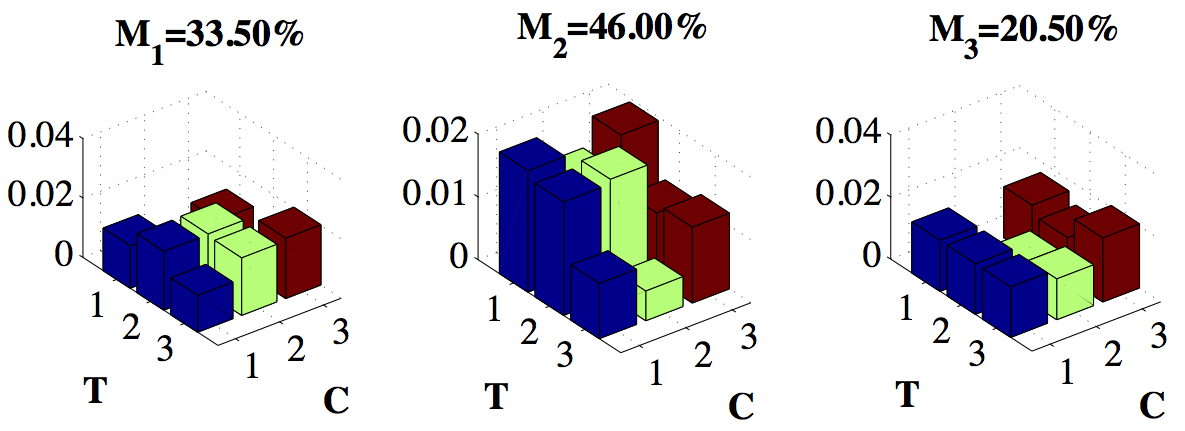}
\end{minipage}}%
\subfloat[Performance for dataset1]{\begin{minipage}[b][6.8cm][t]{.6\textwidth}
\centering
\label{fig-sparsity_clust_hyp1}\includegraphics[width=1\textwidth]{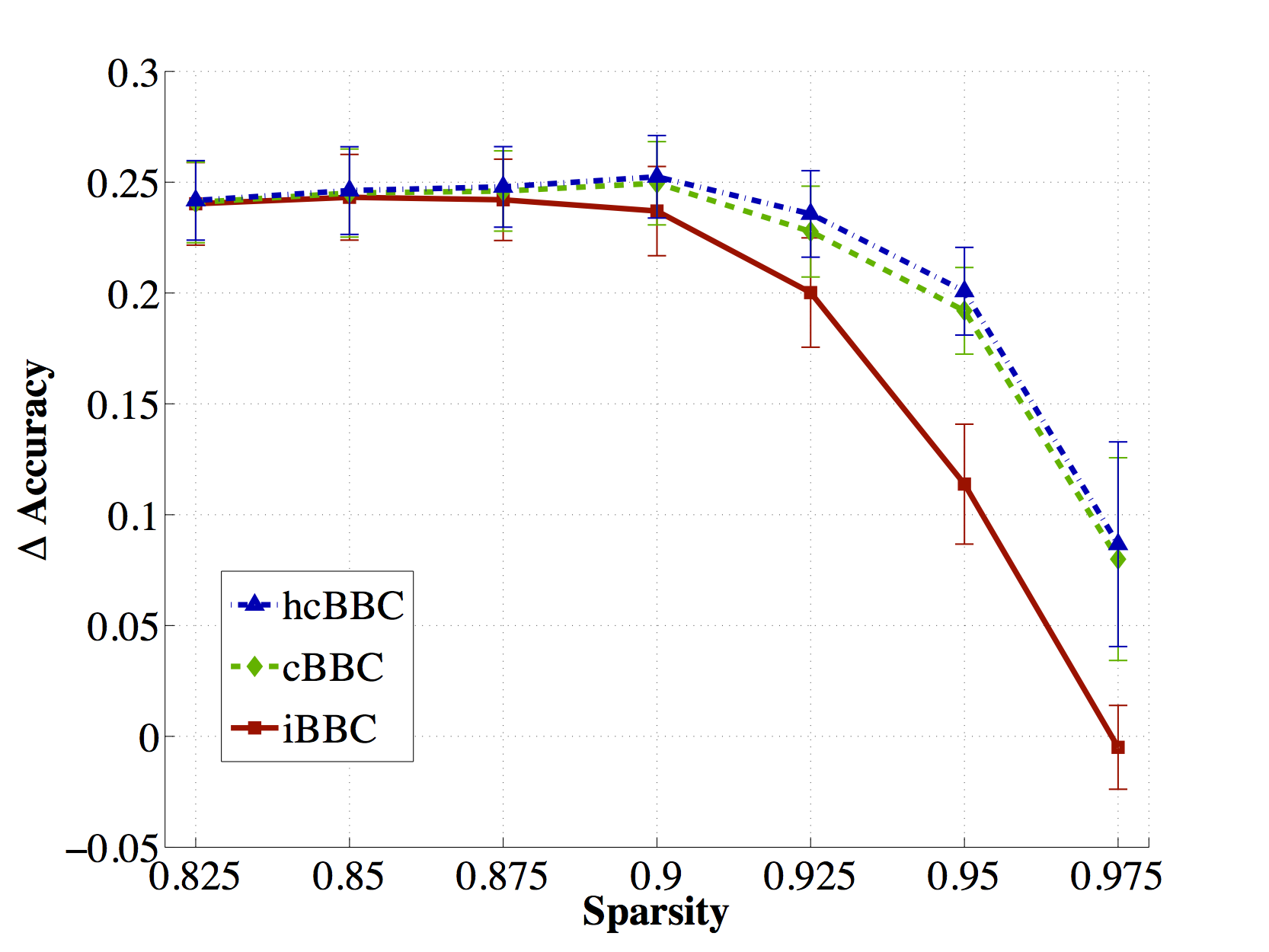}
\end{minipage}}
\caption{Results for dataset1. a) Characteristics of the users' clusters present in dataset1 and the percentage of users in each of them. b) Results for dataset 1. Improvement in accuracy of the different methods with respect to majority voting, for different sparsity levels.}
\end{figure}

Firstly, we generate a synthetic database called dataset1 following the generative model for the hcBCC model. This dataset has 500 labeled instances provided by 200 users. The number of categories is $C=3$. These users belong to 3 clusters with properties shown in Figure \ref{fig-DB_mean}, where we can see the mean of each cluster, their variances and the percentage of users allocated to each of them.

We analyze the behavior of the different algorithms with respect to the sparsity of the input matrix $\bm{Y}$. In particular, we randomly erase a percentage of the entries from $82.5\%$ missing entries to $97.5\%$ in steps of $2.5\%$.  This high sparsity levels are typical in crowdsourcing applications, where the idea is to distribute the load of labeling a dataset among many users, and therefore each of them only labels a small subset of the dataset. 


In the iBCC, the diagonal elements of $\bm{\eta}$ are 0.7 while the off diagonal are 0.3, which reflect our prior belief that users perform better than random.  All the elements of $\bm{\beta}$ are 3. In the cBCC model, the hyper parameters of $\alpha$ are $a_\alpha=1$ and $b_\alpha=10$. This values agree with our prior belief that if the annotations are very scarce, simpler algorithm like majority voting are more suitable and therefore, we should favor partitions with a small number of clusters. For these parameters, in the limiting case when the sparsity of $\bm{Y}$ tends to $100\%$, the average number of clusters tends to 1. Finally, for the hcBCC model we set $\bm{\gamma}$ and $\bm{\phi}$ to the values of $\bm{\eta}$ and $\bm{\beta}$ respectively in the iBCC and cBCC models. All the components of $a_t$ are set to 30 while all the components of $b_t$ are set to 2. This reflects our prior belief that the variability among the users inside the clusters should be less than the variability across clusters.

We run the MCMC  for 10.000 iterations. After the first 3.000  we collect 7.000 samples to compute $\bm{z}$ and $\bm{\pi}$. In the cBCC and hcBCC, we set to five the number of iterations used to sample $\alpha$ following the algorithm proposed in \cite[]{Escobar:1994}. In the hcBCC we fix the number of auxiliary clusters used by the \textit{Reuse Algorithm} to $h=10$.  

%

%
%
%

The increment in accuracy of ours proposals and the iBBC algorithm with respect to majority voting is shown in Figure \ref{fig-sparsity_clust_hyp1}. The two proposed models outperform iBCC as expected.  This improvement of both methods cBCC and hcBCC is particularly significant when the level of sparsity is high, which is a situation that we face in the early stages of a crowdsourcing project. In this case there is not enough information to accurately estimate the confusion matrix of every user independently. We can see that the performance of iBBC drops below the performance of the majority voting algorithm, which assumes all users are similar. Therefore, identifying a clustering structure that allows to share some parameters among the users helps to increase the accuracy of the estimates. Notice that performance of \cite[]{Simpson:11} would be equal to the the iBBC model given that it identifies the users' clusters after the ground truth has been estimated, so it does not affect the performance of the algorithm. 

\begin{figure*}[]
\centering
\begin{tabular}{cc}
\subfloat[cBCC]{{\label{fig-sparsity_clust}}\includegraphics[width=0.95\textwidth]{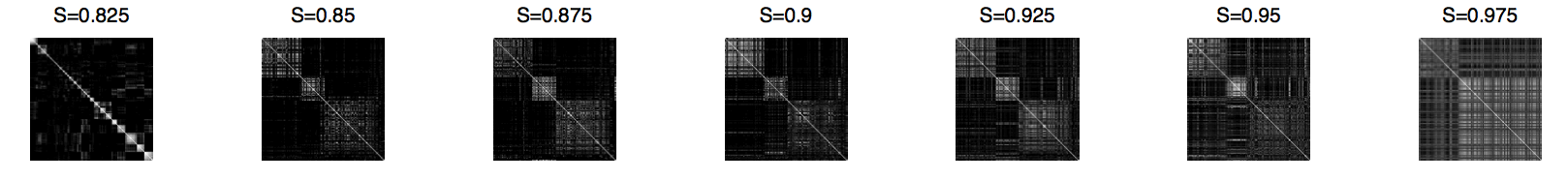}}\vspace{- 3.5mm} &
\multirow{-3}[0]{*}{{{\label{bar1}}\includegraphics[width=0.03\textwidth]{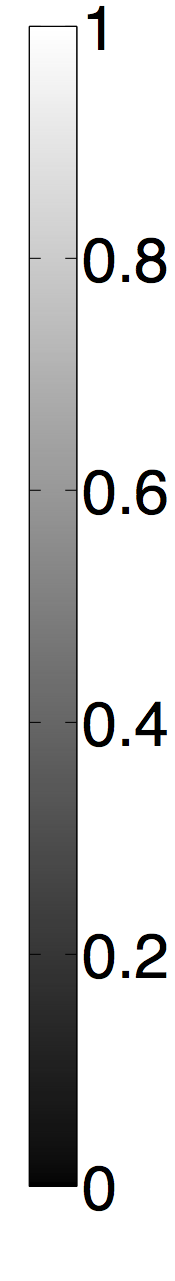}}}\\
\subfloat[hcBCC]{{\label{fig-sparsity_normal}}\includegraphics[width=0.95\textwidth]{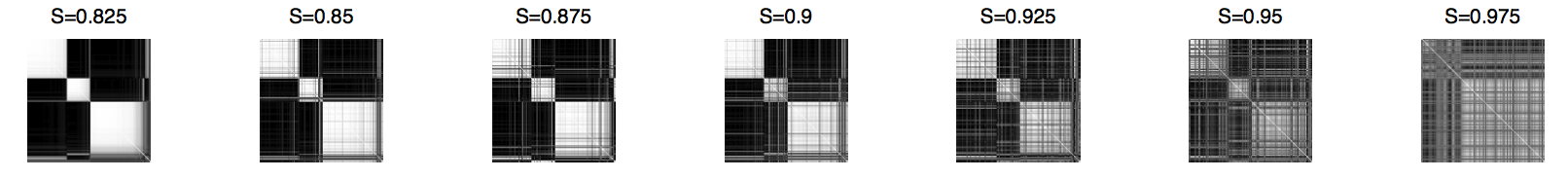}}
\end{tabular}
\caption{Co-occurency matrices of the users}
\label{results-clust-hyp-clusters}
\vspace{0mm}
\end{figure*}

To further analyze the cluster structure identified by the algorithms, in Figure \ref{results-clust-hyp-clusters} we represent the co-ocurrence matrix of the users. The position $(\ell,\ell')$ is the probability of $\ell$ and $\ell'$ belonging to the same cluster. We can see that the clusters identified by the hcBCC are more useful than the ones extracted by the cBCC, because in a practical situation we are not normally interested in finding users with exactly the same behavior, but users with similar characteristics. For example, we can see that when $82.5\%$ of the annotations are missing, the hcBCC algorithm identifies the 3 main groups of users while the cBCC algorithm identifies instead a much larger number of groups because of the constraint that all users of a cluster must have the same properties. So, although both algorithms' performance is similar, the clustering provided by the hBCC is easier to interpret and gives a simpler explanation of the data. 


\begin{figure}[t]
\vspace{0mm}
\centering
\vspace{0mm}
\begin{tabular}{c}
\subfloat[Performance for dataset2]{{\label{fig-sparsity_clust_results}}\includegraphics[width=0.52\textwidth]{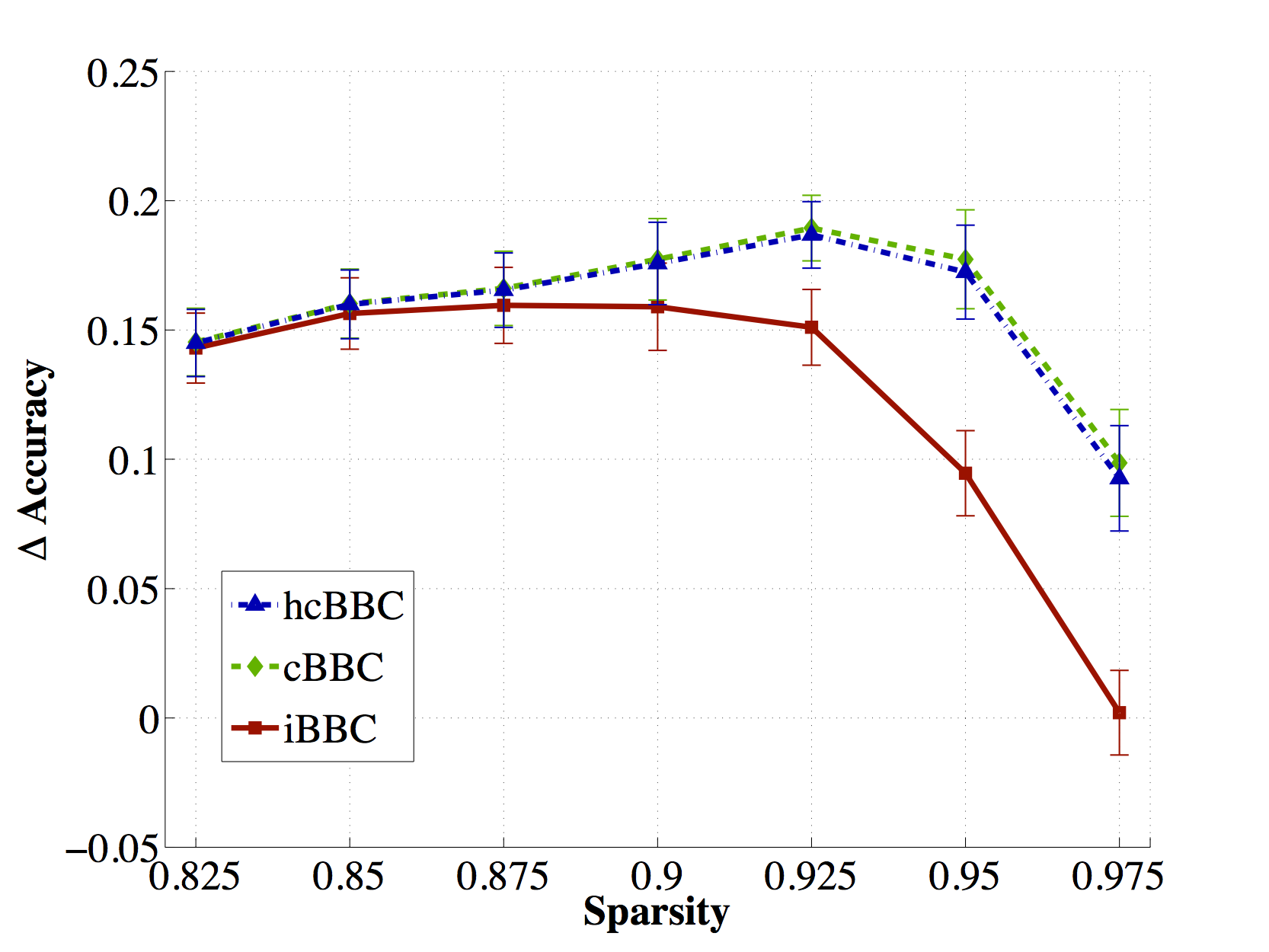}} 
\subfloat[Performance for dataset3]{{\label{fig-sparsity_normal_results}}\includegraphics[width=0.52\textwidth]{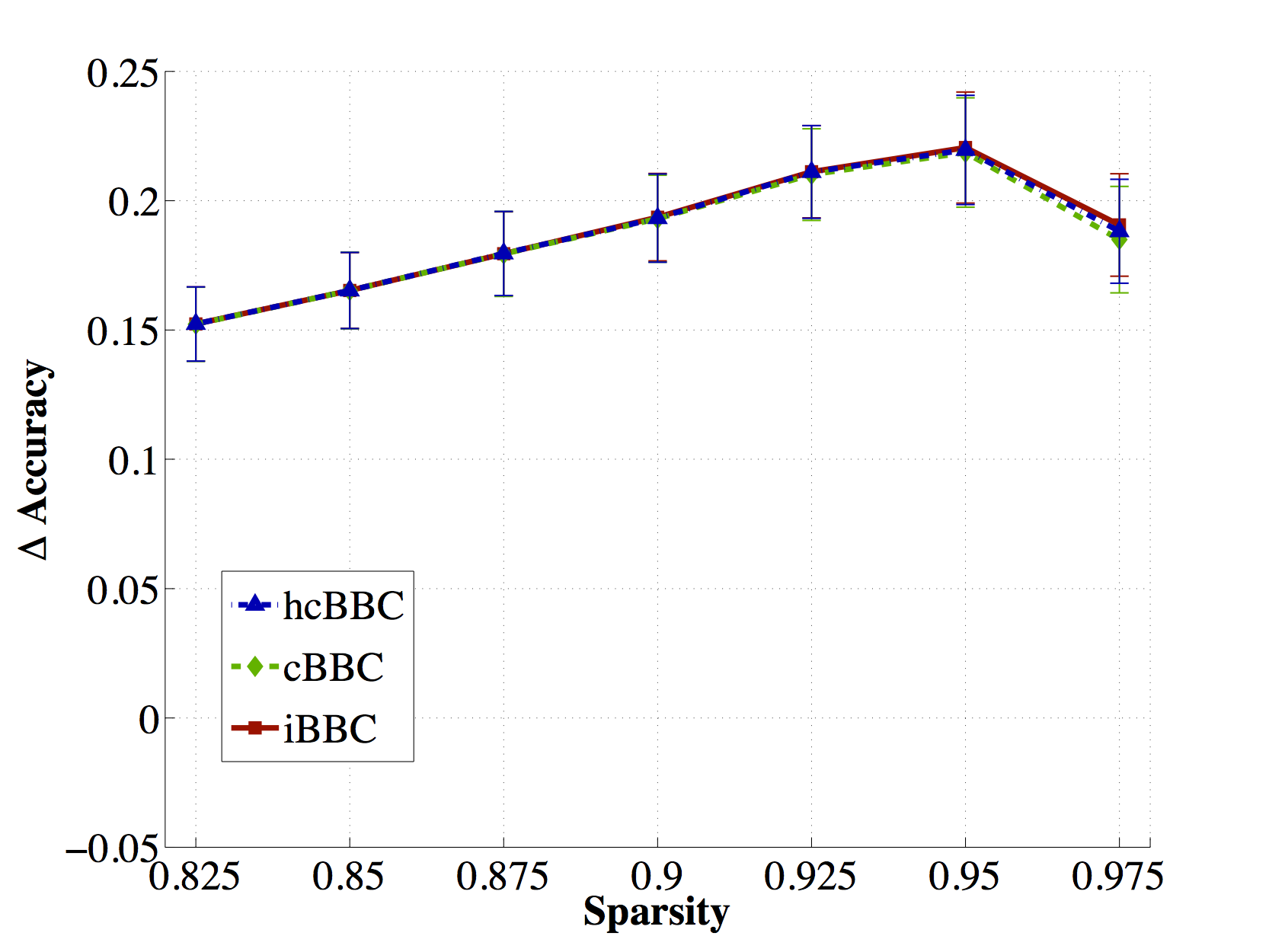}}
\end{tabular}
\caption{Results for dataset2 and dataset3. Improvement in accuracy of the different methods with respect to majority voting, for different sparsity levels}
\label{datasets23}
\end{figure}

Finally, we test with datasets that are generated following the iBCC and cBCC models. First, we create a new dataset (dataset2) in which the mean confusion matrix of each cluster is the same as in dataset1 which is shown in Figure \ref{fig-DB_mean}. However, in this case the variability of the confusion matrices inside each cluster is zero. Therefore, this new dataset follows the assumptions made by the cBCC model. Again, the performance of the cBCC and the hcBCC models outperforms iBCC as expected (see Figure \ref{fig-sparsity_clust_results}). However, even though data is generated from the cBCC which is a simpler model than the hcBCC, hcBCC is able to discover the underlying structure of the users and gets a performance which is on par with the cBCC. The hcBCC does not degrade the solution although it is more flexible. 

%

In the last database called dataset3, we generate all the instances from the same clustering ($M_2$ in Figure \ref{fig-DB_mean}). In this case there is no different clusters of users and each of them has its own confusion matrix. Therefore, this dataset fulfill the assumptions of the iBCC. In Figure \ref{fig-sparsity_normal_results} we see that the performance of the two proposed models is identical to iBCC. To sum up, we see that the performances of cBCC and hcBCC dominate iBCC under all conditions tested.

\subsection{Real datasets}

In this Section, we use $4$ publicly available crowdsourced datasets with $C=2$ whose principal characteristics are described in Table \ref{tab-real_properties_dbs} \cite[]{Raykar:2012}. 

To choose the hyper-parameters we follow the reasoning of Section \ref{synthetic}. Specifically, in the iBCC the diagonal elements of $\bm{\eta}$ are 0.7 while the off diagonal are 0.3, and all the elements of $\bm{\beta}$ are set to 3. In this way, we incorporate our prior belief that users are imperfect but perform better than chance. In the cBCC model we use the same value for $\bm{\eta}$ and $\bm{\beta}$ so that the comparison is fair. Finally for the hcBCC model, $\bm{\gamma}$ is set to the same value used for  $\bm{\eta}$ in the previous models. All the components of $a_t$ are set to 20 while all the components of $b_t$ are set to 2, reflecting our belief that the variability inside clusters should be lower than the variability across clusters. We fix $a_\alpha=1,b_\alpha=10$ in both, cBCC and hcBCC.  We run the MCMC for 10.000 iterations and we discard the first 3.000 to compute the posterior distribution of $\bm{z}$ and $\bm{\pi}$.

 In Table \ref{acc_real}, we see the performance of the different algorithms in terms of accuracy predicting the ground truth. In particular, we see that the performances of the cBCC and the hcBCC are better than that of the iBCC in the last three datasets, i.e. rte, temp and valence. On the other hand, in the bluebird datasets the iBBC performs better. Notice again that the performance of the algorithm described in \cite[]{Simpson:11} would be exactly equal to the one of the iBCC, given that the communities of users are inferred after the ground truth is inferred and therefore, it does not affect the accuracy in any way. 

\begin{figure}[]
\centering
\begin{tabular}{c}
\subfloat[bluebird]{{\label{fig-bluebird}}\includegraphics[width=0.17\textwidth]{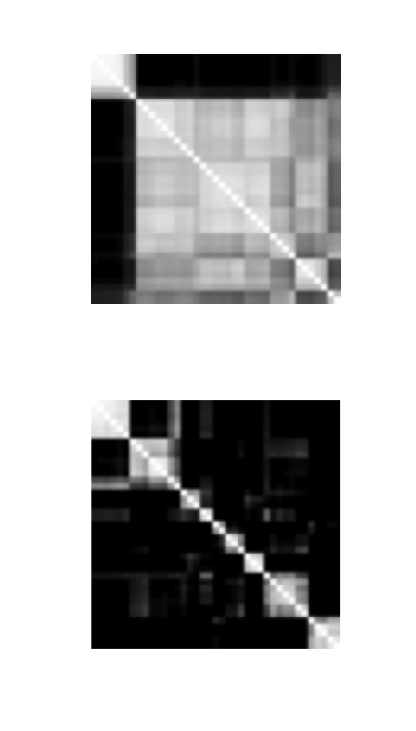}}
\subfloat[rte]{{\label{fig-rte}}\includegraphics[width=0.17\textwidth]{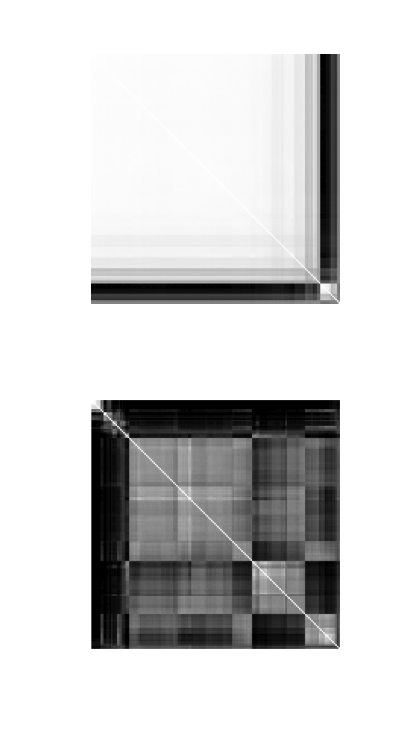}}
\subfloat[valence]{{\label{fig-valence}}\includegraphics[width=0.17\textwidth]{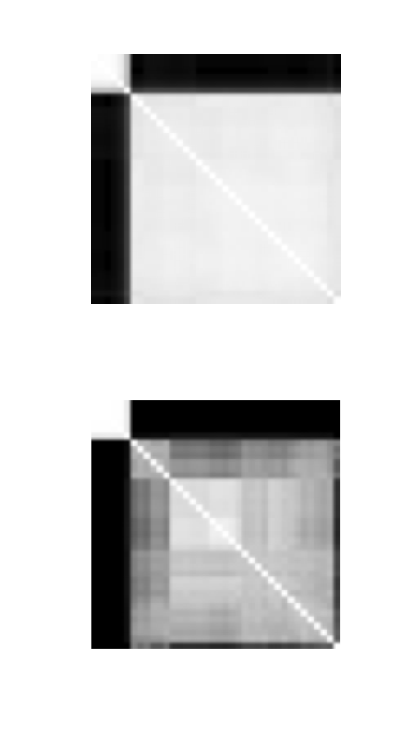}}
\subfloat[temp]{{\label{fig-temp}}\includegraphics[width=0.17\textwidth]{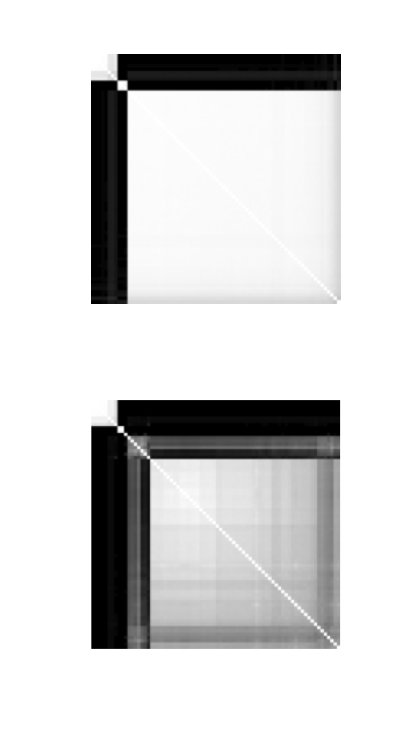}}\\
\end{tabular}
\caption{Co-ocurrency matrix of the users. (Upper row) hcBCC. (Lower row) cBCC.}
\label{clustreal}
\end{figure}

The performance difference between the cBCC and the hcBCC is only significant in the valence dataset. However, the main advantage of the hcBCC model over the cBCC is clear when we represent the average number of clusters (See Figure \ref{clustreal} and Table \ref{acc_real}). Even though the cBBC model correctly captures the clustering structure of the users, forcing all users of a cluster to share the same confusion matrix translates into a large number of clusters, some of them with very similar properties. 

The hcBCC identifies a smaller number of clusters that are much more interpretable, in the sense that it perfectly identifies each kind of clusters thanks to its additional flexibility. We are not interested in identifying clusters of users with the exact same behavior, but what we really want is to find clusters of users that behave in a similar way, so we can establish strategies to boost the overall performance of the crowdsourcing system, e.g. by rewarding the most efficient labelers, avoiding spammers or by better defining the description of the task based on the biases identified in the clusters of users.  



\begin{table}[]
\renewcommand{\tabcolsep}{1mm}
\centering
\begin{tabular}{lcccccp{8cm}}
\hline
 Dataset & N & L & $\mu_n$ & $\mu_l$ &  Sparsity (\%) & Brief Description\\
 
\hline
bluebird & 108 & 39 & 108 & 39  & 0 & \begin{scriptsize}Identify wether there is a Indigo Bunting or Blue Grosbeak in the image\end{scriptsize}\\
rte & 800 & 164 & 49 & 10 & 93.90 &  \begin{scriptsize}Identify wether the second sentence can be inferred from the first\end{scriptsize}\\
valence & 100 & 38 & 26 & 10 & 73.68 & \begin{scriptsize}Identify the overall positive or negative valence of the emotional content of a headline\end{scriptsize}\\
temp & 462 & 76 & 61 & 10 & 86.84 & \begin{scriptsize}Users observe a dialogue and two verbs from the dialogue and have to identify whether the first verb event occurs before or after the second \end{scriptsize} \\
\hline
\end{tabular}
\caption{ Description of the real datasets. $N$  and $L$ denotes de number of instances and users respectively. $\mu_n$ stand for the mean number of instances labelled by a user and $\mu_l$ designate the mean number of users that label an instance.}
\label{tab-real_properties_dbs}
\end{table}

\begin{table}[]
\renewcommand{\tabcolsep}{1mm}
\centering
\begin{tabular}{lcccccc}
\hline
 Dataset & \multicolumn{4}{ c }{Accuracy(\%)} & \multicolumn{2}{ c }{Average number of clusters}\\
  & Majority & iBCC & cBCC & hcBCC &  cBCC & hcBCC\\
\hline
bluebird & 75.93 & \textbf{89.81} & 88.89  & 88.89 & 11.32 $\pm$ 0.04 & 3.31 $\pm$ 0.09\\
rte & 91.88 &  92.88 & \textbf{93.12}  & \textbf{93.12} & 7.70 $\pm$ 0.07 & 2.30 $\pm$ 0.06\\
valence & 80.00 &  85.00 & 88.00  & \textbf{89.00} & 3.5 $\pm$ 0.04 & 2.25 $\pm$ 0.02\\
temp & 93.94 &  94.35  & \textbf{94.37}  & \textbf{94.37} & 6.20 $\pm$ 0.03 & 3.2 $\pm$ 0.02\\
\hline
\end{tabular}
\caption[aaa]{Results for the real data. Mean accuracy of the different algorithms \footnotemark. Average number of clusters (mean $\pm$ one standard deviation).}
\label{acc_real}
\end{table}

\footnotetext{The standard deviation are less than $10^-4$ and are is not shown}



\begin{figure}[]
\centering
\begin{tabular}{c}
\subfloat[bluebird]{{\label{fig-bluebird_ex}}\includegraphics[width=0.3\textwidth]{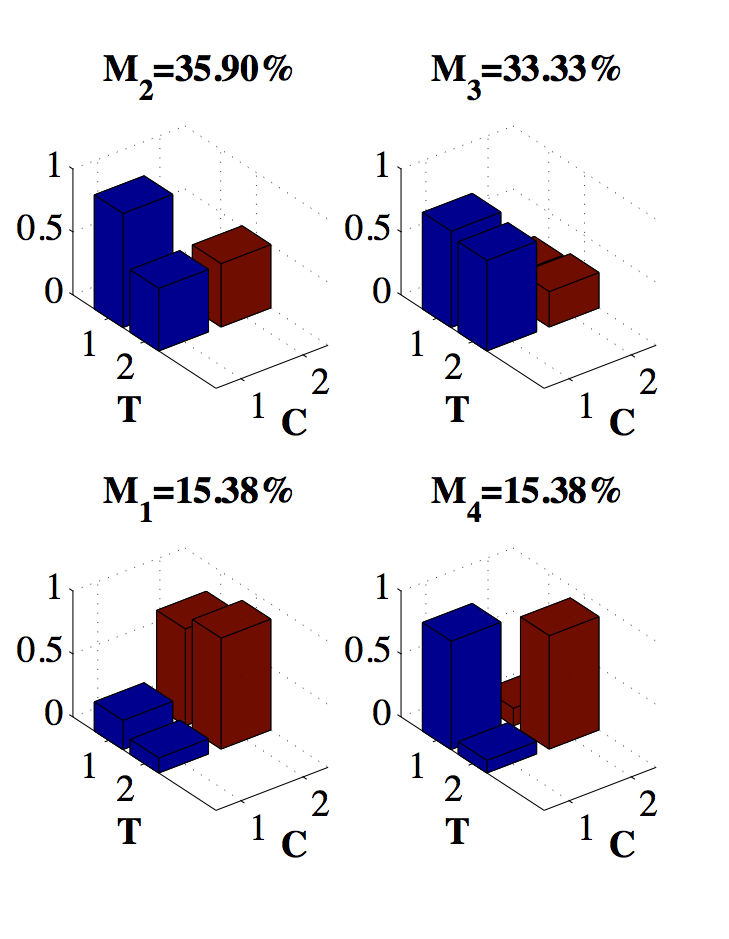}}
\subfloat[temp]{{\label{fig-temp_ex}}\includegraphics[width=0.3\textwidth]{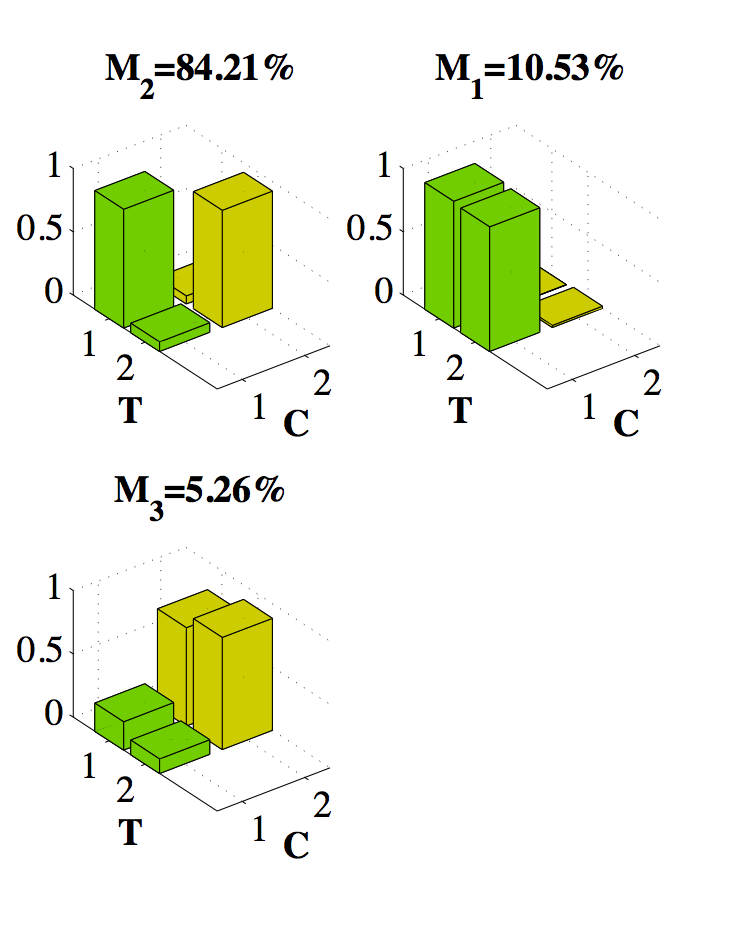}}
\subfloat[rte]{{\label{fig-rte_ex}}\includegraphics[width=0.15\textwidth]{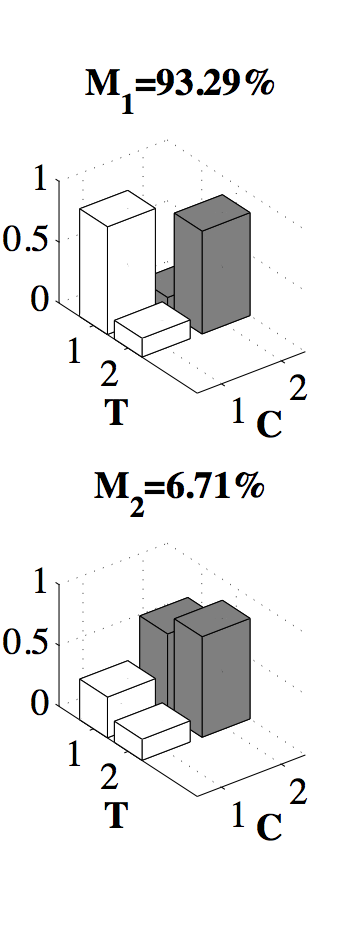}}
\subfloat[valence]{{\label{fig-valence_ex}}\includegraphics[width=0.15\textwidth]{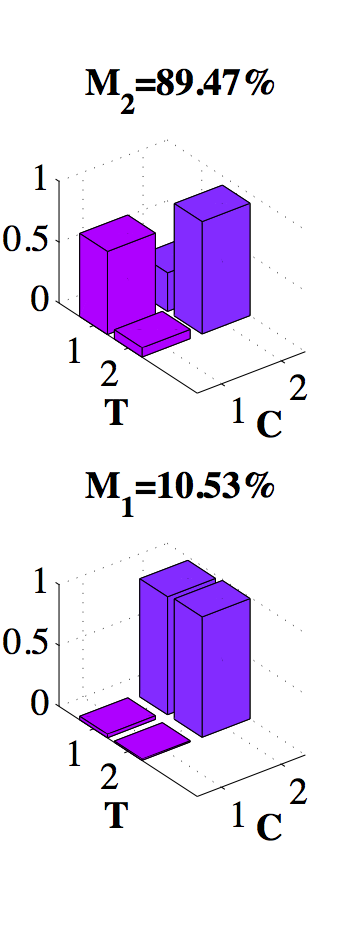}}
\end{tabular}
\caption{Mean confusion matrices of the user's clusters identified by hcBCC}
\label{real_ex}
\end{figure}

In Figure \ref{real_ex} we show as an example the mean confusion matrix of the hcBCC clusters in the datasets. It shows very interpretable clusters that are useful for the modeler. In the bluebird dataset we can clearly identify a small subset of experts ($M_4 = 15.38\%$) who shows a high performance labeling the bird images. In addition, we find that the biggest cluster ($M_2 = 35.90\%$) corresponds to users whose accuracy is high when the real class is  $z=1$ (images of Blue Grosbeak) but performs poorly when the class is $z=2$ (images of Indigo Bunting).  Finally, we have two clusters of spammers. In the first cluster ($M_1 = 15.38\%$)   users tend to label all images as belonging to class $z=2$ and in the second ($M_3 = 33.33\%$)  users tend to label all images as  $z=1$. In the temp dataset, we can observe that the majority of the users ($M_2 = 84.21\%$) are experts, but there are again two small clusters of spammers. As for the rte dataset, most of the users have a good performance ($M_1=93.29\%$). The remaining users are bias toward labeling instances as belonging to class $z=2$. Finally, in the valence dataset we can see that the majority of the users ($M_2=89.47\%$) are very accurate identifying instances belonging to class $z=2$ and have a medium performance when $z=1$. In addition we find a small cluster of users that have labelled almost every instance as $z=2$.  All this information about the underlying clustering structure of the users in the datasets can be used in a real crowdsourcing application  to develop efficient strategies to minimize the  cost of a crowdsourcing project maximizing the performance.

\begin{figure}[t]
\vspace{0mm}
\centering
\vspace{0mm}
\begin{tabular}{c}
\subfloat[Performance for bluebird]{{\label{fig2-sparsity_clust_results}}\includegraphics[width=0.5\textwidth]{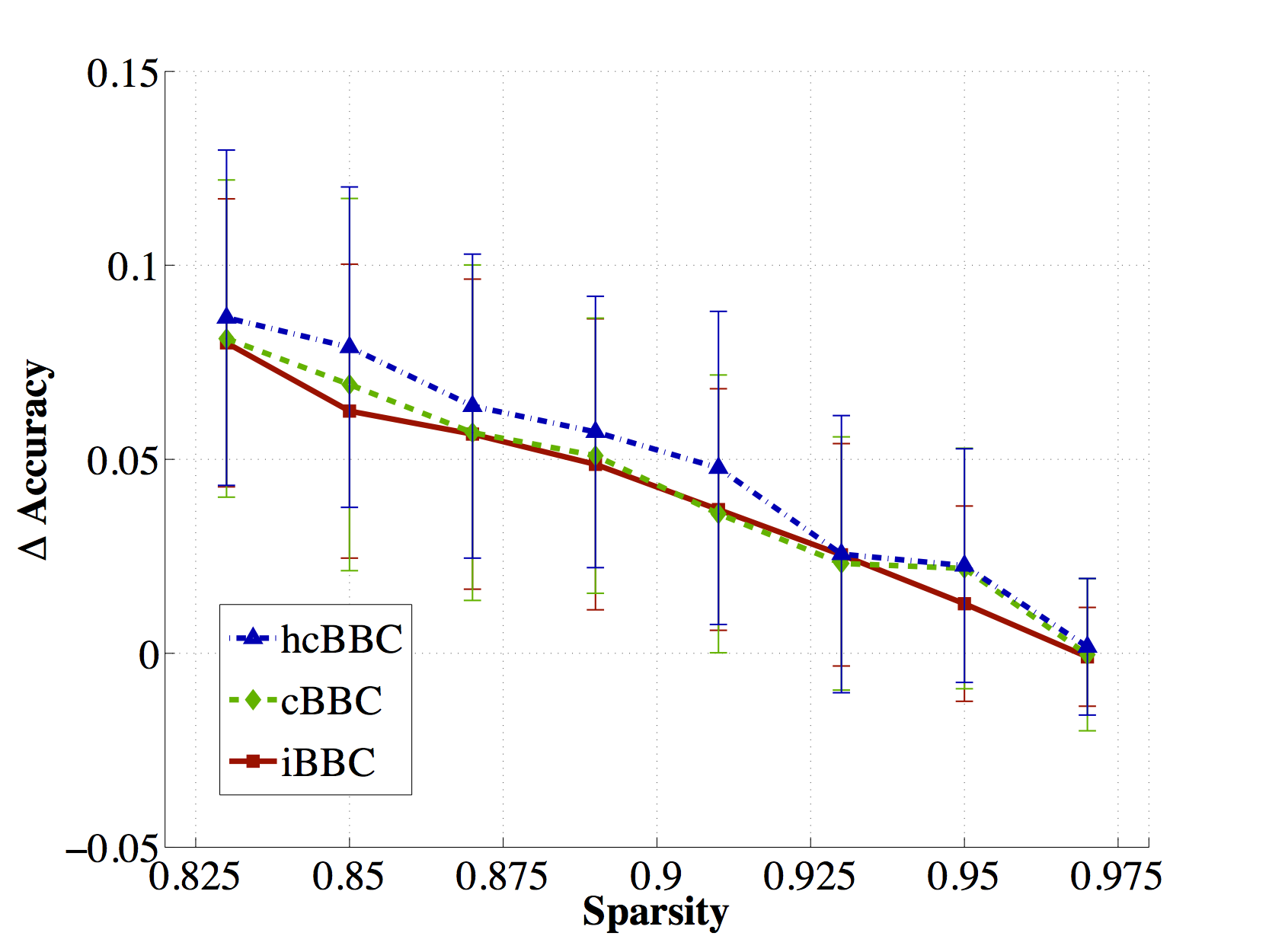}} 
\subfloat[Performance for rte]{{\label{fig2-sparsity_normal_results}}\includegraphics[width=0.5\textwidth]{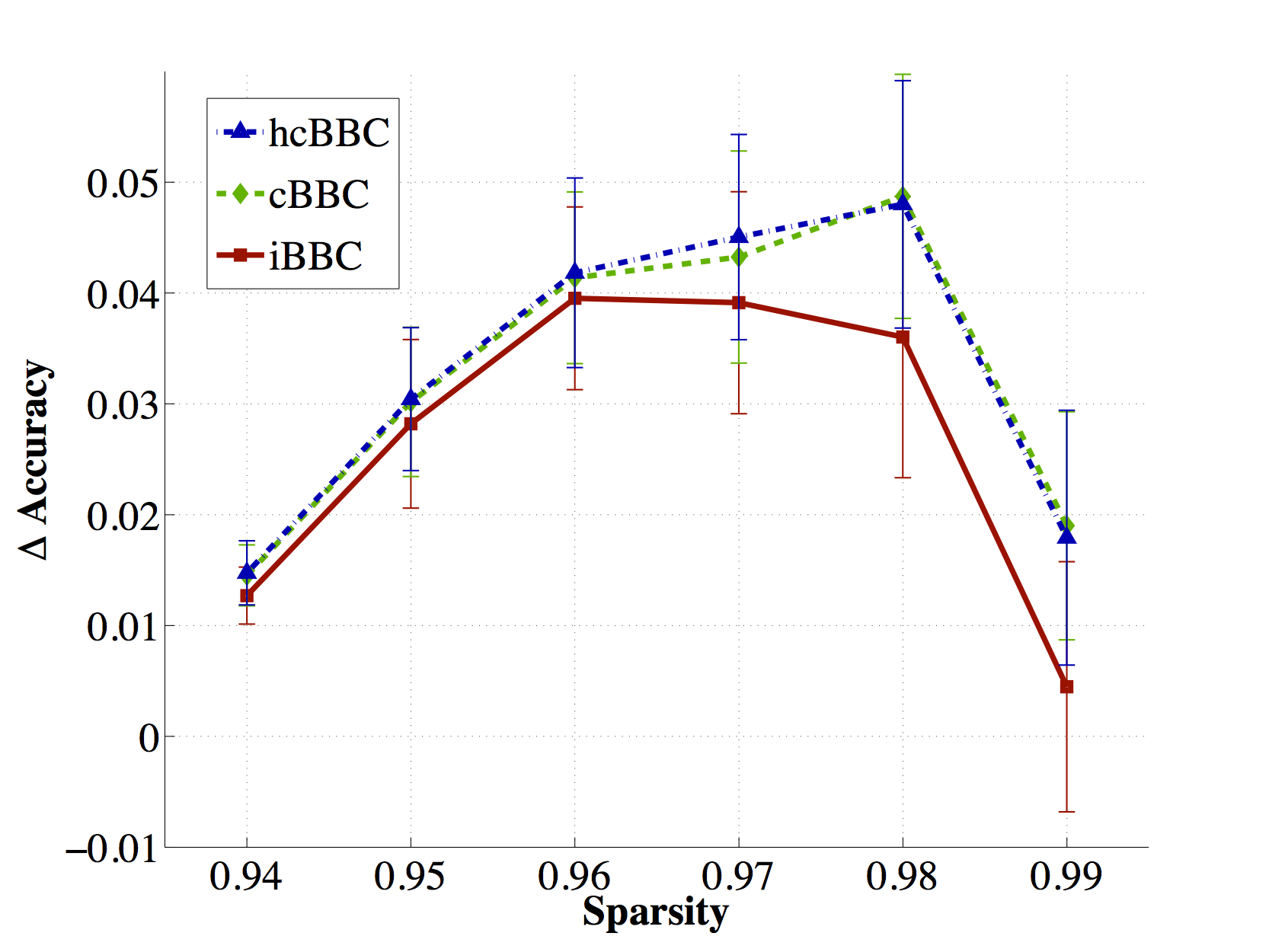}}\\
\subfloat[Performance for valence]{{\label{fig2-sparsity_clust_results}}\includegraphics[width=0.5\textwidth]{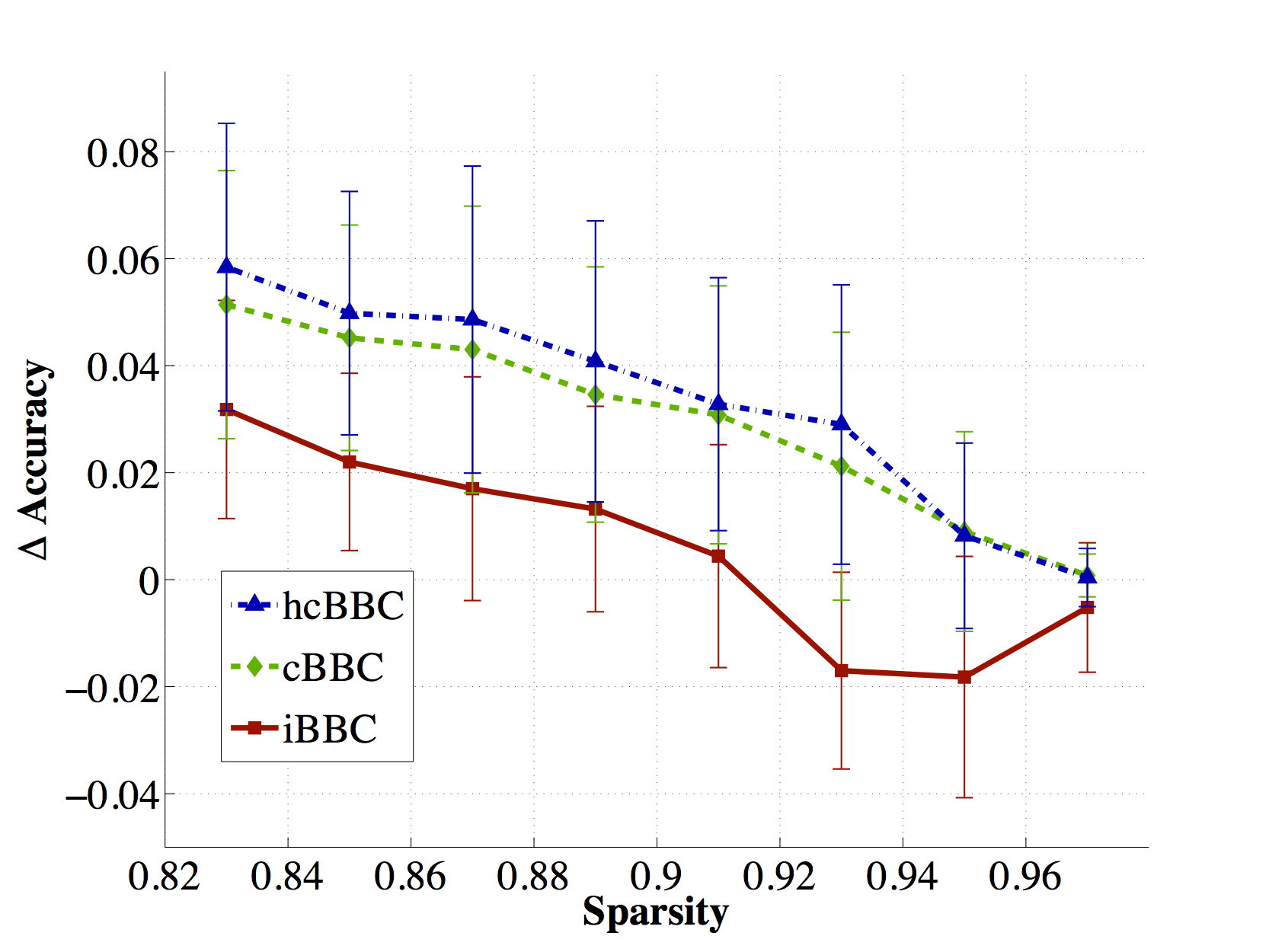}} 
\subfloat[Performance for temp]{{\label{fig2-sparsity_normal_results}}\includegraphics[width=0.5\textwidth]{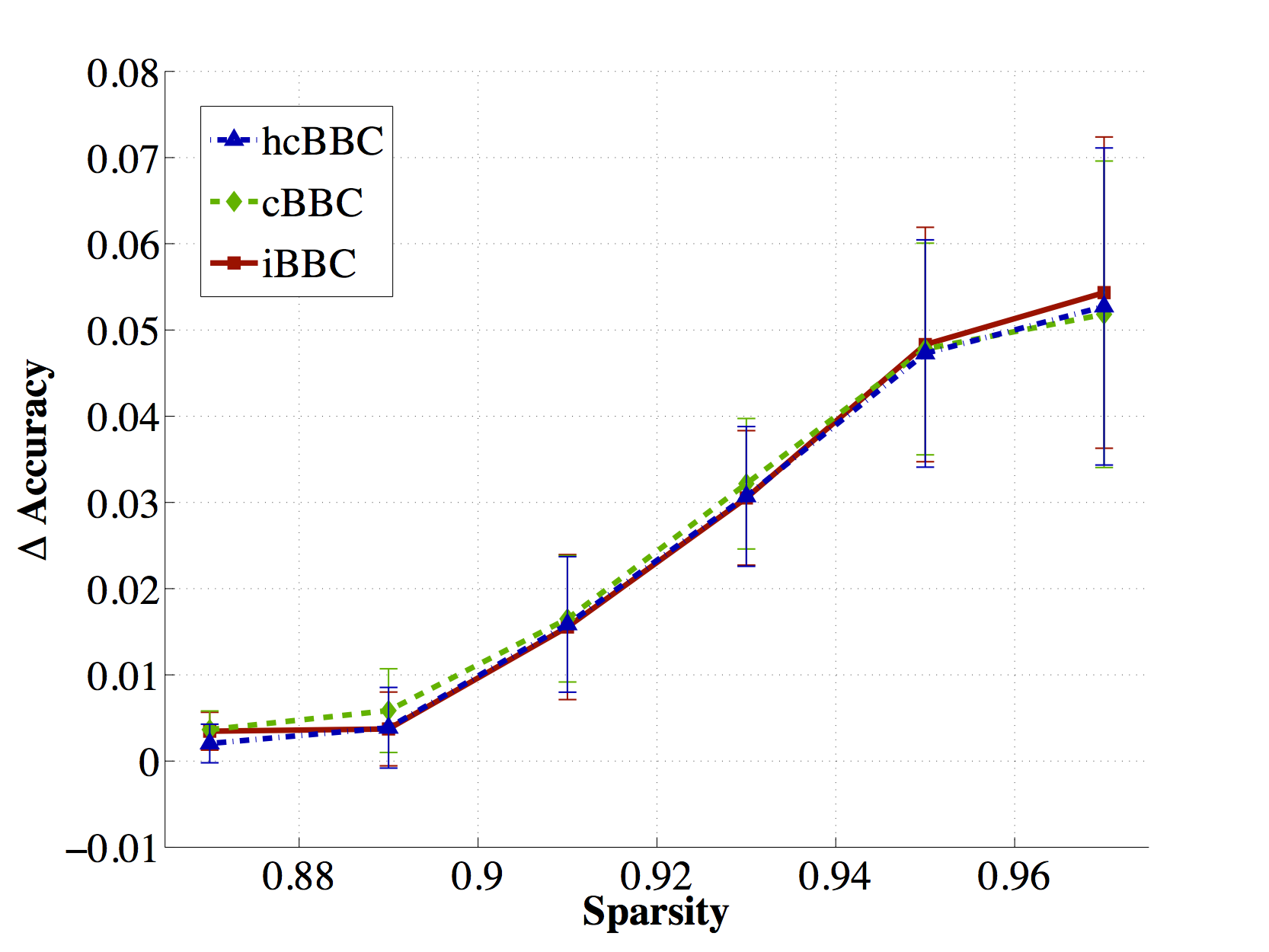}}
\end{tabular}
\caption{Results for real datasets. Improvement in accuracy of the different methods with respect to majority voting, for different sparsity levels}
\label{real_sparsity}
\end{figure}

To conclude this Section, we evaluate the performance of the algorithms in the real datasets for different levels of sparsity. Following the procedure in Section \ref{synthetic}, we create 50 random databases for each level of sparsity. We do that in such a way that every instance has at least one label and every user provides at least one label. The results are shown in Figure \ref{real_sparsity}.

In the dataset bluebird and temp, we observe that finding clusters of users does not have a significant effect in terms of accuracy. However, the cBCC and the hcBCC models do not degrade the performance and give us some insight about the users in the crowdsourcing application (See Figure \ref{real_ex}). In the rte and valence datasets, the inference of the clustering structure of the users also translates into an improvement in terms of accuracy. In the rte dataset, this improvement is not significant for the original sparsity level, but it becomes more significant when the sparsity is increased. What happens is that when the sparsity is very high, there are very few annotations provided by each user, and the iBBC algorithm fails to infer the properties of each user separately. 

In the valence dataset, we can even see that the performance of the iBBC model drops below the performance of a simple majority voting algorithm when the sparsity is increased. However, the cBCC and hcBCC outperform the majority voting algorithm for every sparsity level. Again the iBBC model does not have enough information to infer the properties of each user and a simpler model like majority voting, which assume that all users have the same level of expertise, performs better. Actually, what is happening is that the the CRP prior used in the cBCC and the hcBCC models favors partitions with a small number of clusters. When the input matrix $\bm{Y}$ is very sparse, the prior term dominates over the likelihood and all users tend to be grouped in the same cluster.




\section{Conclusions}
\label{Conclusions}

We have proposed two new Bayesian nonparametric models to merge the information provided by the users in a crowdsourcing system. In addition, the algorithms detect clusters of users that have similar behaviors and use this information to improve the ground truth estimate. In the cBCC model, we have used a CRP to infer the partitioning of the users such that users in the same cluster are constrained to have the same properties. In the hcBCC model, we have used a hierarchical structure to increase the flexibility. In particular, each user has its own properties, but users assigned to the same cluster have similar properties. In this way, it finds smaller number of clusters that are easy to interpret. 

We have shown how these new models relate to the iBCC model and analyzed the correlation structure among the users as a consequence of the clustering. We have proposed MCMC methods to infer the parameters of both models and performed several experiments with synthetic and real databases, which have shown that the algorithms outperform the current state-of-the-art.

Finally, we comment possible extensions. The ground truth estimated by the proposed algorithms,  can be used to train a supervised learning algorithm. \cite[]{Raykar:2010,Groot11,Wellinder10} propose to train a classifier directly with the noisy labels provided by the users. It would be interesting to extend the models following this line. Also, the models assume consistent users. An extension would be considering users with nonuniform behavior \cite[]{Zhang:2011,Yan:2010}, i.e.  a user can be an expert for a subset of the instances while can act as a novice in another subset.


\acks{Pablo G. Moreno is supported by an FPU fellowship from the Spanish Ministry of Education
(AP2009-1513). This work has been partly supported by Ministerio de  Economía of Spain ('COMONSENS', id. CSD2008-00010, 'ALCIT', id. TEC2012-38800-C03-01, 'COMPREHENSION', id. TEC2012-38883-C02-01). This work was also supported by the European Union 7th Framework Programme through the Marie Curie Initial Training Network "Machine Learning for Personalized Medicine" MLPM2012, Grant No. 316861. Yee Why Teh's research leading to these results has received funding from the European Research Council under the European Union's Seventh Framework Programme (FP7/2007-2013) ERC grant agreement no. 617411.}



\appendix
\section*{Appendix A.}
\label{app:correlation}

In the iBCC model, the joint probability of two users given the ground truth is:

\begin{equation}
\begin{split}
&p(y_{i\ell}, y_{i\ell'} | z_i=t) = \frac{\Gamma(\beta_t)}{\Gamma(\beta_t + \mathds{I}(y_{i\ell} \neq 0))} \prod_c \frac{\Gamma(\beta_t \eta_{tc} + \mathds{I}(y_{i\ell} = c, y_{i\ell} \neq 0))}{\Gamma(\beta_t \eta_{tc})} \times\\ &\times \frac{\Gamma(\beta_t)}{\Gamma(\beta_t + \mathds{I}(y_{i\ell'} \neq 0))} \prod_c \frac{\Gamma(\beta_t \eta_{tc} + \mathds{I}(y_{i\ell'} = c, y_{i\ell'} \neq 0))}{\Gamma(\beta_t \eta_{tc})} = p(y_{i\ell} | z_i=t) \times p(y_{i\ell'} | z_i=t) 
\end{split}
\end{equation}

Therefore:

\begin{equation}
\mathrm{corr}(\mathds{I}(y_{i\ell}=a),\mathds{I}(y_{i\ell'}=b)) = 0
\end{equation}

In the cBCC model, we have the following expression for the joint distribution of $y_{i\ell}$ and $y_{i\ell'}$:

\begin{equation}
{\begin{split}
&p(y_{i\ell}, y_{i\ell'} | z_i=t) = \left( \frac{1}{1 + \alpha} \right) \left[ \frac{\Gamma(\beta_t)}{\Gamma(\beta_t + \mathds{I}(y_{i\ell} \neq 0) + \mathds{I}(y_{i\ell'} \neq 0))} \times \right. \\
&\left. \times \prod_c \frac{\Gamma(\beta_t \eta_{tc} + \mathds{I}(y_{i\ell} = c, y_{i\ell} \neq 0) + \mathds{I}(y_{i\ell'} = c, y_{i\ell'} \neq 0))}{\Gamma(\beta_t \eta_{tc})}\right] + \left( \frac{\alpha}{1 + \alpha} \right) \left[ \frac{\Gamma(\beta_t)}{\Gamma(\beta_t + \mathds{I}(y_{i\ell} \neq 0))} \times \right.\\
&\left. \prod_c \frac{\Gamma(\beta_t \eta_{tc} + \mathds{I}(y_{i\ell} = c, y_{i\ell} \neq 0))}{\Gamma(\beta_t \eta_{tc})} \times \frac{\Gamma(\beta_t)}{\Gamma(\beta_t + \mathds{I}(y_{i\ell'} \neq 0))} \prod_c \frac{\Gamma(\beta_t \eta_{tc} + \mathds{I}(y_{i\ell'} = c, y_{i\ell'} \neq 0))}{\Gamma(\beta_t \eta_{tc})} \right]
\end{split}}
\end{equation}

We can now compute the covariance in the following way:

\begin{equation}
{\begin{split}
&\mathrm{cov}(\mathds{I}(y_{i\ell}=a),\mathds{I}(y_{i\ell'}=b)) = \mathds{E} \{ \mathds{I}(y_{i\ell}=a) \mathds{I}(y_{i\ell'}=b)  \} - \mathds{E} \{ \mathds{I}(y_{i\ell}=a)  \}\mathds{E} \{ \mathds{I}(y_{i\ell'}=b)  \} =\\
&= \left( \frac{1}{1 + \alpha} \right) \left[ \frac{\Gamma(\beta_t)}{\Gamma(\beta_t + \mathds{I}(a \neq 0) + \mathds{I}(b \neq 0))} \prod_c \frac{\Gamma(\beta_t \eta_{tc} + \mathds{I}(a = c, a \neq 0) + \mathds{I}(b = c, b \neq 0))}{\Gamma(\beta_t \eta_{tc})} - \right.\\
&\left. - \frac{\Gamma(\beta_t)}{\Gamma(\beta_t + \mathds{I}(a \neq 0))} \prod_c \frac{\Gamma(\beta_t \eta_{tc} + \mathds{I}(a = c, a \neq 0))}{\Gamma(\beta_t \eta_{tc})} \times \frac{\Gamma(\beta_t)}{\Gamma(\beta_t + \mathds{I}(b \neq 0))} \prod_c \frac{\Gamma(\beta_t \eta_{tc} + \mathds{I}(b = c, b \neq 0))}{\Gamma(\beta_t \eta_{tc})}\right]
\end{split}}
\end{equation}

Assuming that $a \neq 0$ and $b \neq 0$, and considering the cases where $a =b$ and $a \neq b$ we obtain the following equation for the covariance:

 \begin{equation}
\mathrm{Cov}(\mathds{I}(y_{i\ell}=a),\mathds{I}(y_{i'\ell'}=b) | z_i = t)  =
\begin{cases}
 - \left( \frac{1}{1+\alpha} \right) \left(   \frac{1}{1 + \beta_t} \right) \eta_{ta} \eta_{tb} \hspace{5mm} a \neq b\\
 \left( \frac{1}{1+\alpha} \right) \left(   \frac{1}{1 + \beta_t} \right) \eta_{ta} (1 - \eta_{ta}) \hspace{5mm} a = b
 \end{cases}
\end{equation}

Here we have taken into account that $\Gamma(x+1)=x\Gamma(x)$. Once we get the expression of the covariance, we divide it by the square root of the variances to get the correlation:

\begin{equation}
\mathrm{Corr}(\mathds{I}(y_{i\ell}=a),\mathds{I}(y_{i\ell'}=b)) = \frac{\mathrm{Cov}(\mathds{I}(y_{i\ell}=a),\mathds{I}(y_{i\ell'}=b))}{\sqrt{\mathrm{Var}(\mathds{I}(y_{i\ell}=a))\mathrm{Var}(\mathds{I}(y_{i\ell'}=b))}}
\end{equation}

It is straightforward to see that $\mathrm{Var}(\mathds{I}(y_{i\ell}=a)) = \eta_a (1 - \eta_a)$, getting the expected result.

\section*{Appendix B.}

\label{app:posterior}

The posterior distribution of the parameters $\bm{\beta}^m$ and $\bm{\eta}^m$  is proportional to the following expression:

\begin{equation}
{\begin{split}
&p(\bm{\eta}, \bm{\beta} | \bm{Y}, \bm{z}, \bm{\pi}) \propto p(\bm{Y} |\bm{\eta}, \bm{\beta}, \bm{z}, \bm{\pi}) \times p(\bm{\eta}, \bm{\beta}) = \prod_\ell \prod_t  \left[ \frac{\Gamma(\beta_t^{q_\ell})}{\Gamma(n_{\ell t} + \beta_t^{q_\ell})} \prod_c \frac{\Gamma(n_{\ell tc} + \beta_t^{q_\ell} \eta_{tc}^{q_\ell})}{\Gamma(\beta_t \eta_{tc}^{q_\ell})} \right] \times \\
&\times \prod_m \prod_t \left[ \frac{\Gamma(\phi_t)}{\prod_c \Gamma(\phi_t \gamma_{tc})} \prod_c \left( \eta_{tc}^m \right) ^{\phi_t \gamma_{tc} - 1} \right]  \prod_m \prod_t \frac{b_t^{a_t}}{\Gamma(a_t)} (\beta_t^m)^{a_t - 1} \exp (-b_t \beta_t^m) \\
\end{split}}
\end{equation}

We cannot compute an analytic expression for $p(\bm{\eta}, \bm{\beta} | \bm{Y}, \bm{z}, \bm{\pi})$ because the prior on $p(\bm{\eta}, \bm{\beta})$ is no longer conjugate of the likelihood of the observations. The idea is to include two auxiliary variables $\bm{\nu}$ and $\bm{s}$ such that we can compute the  joint distribution $p(\bm{\eta}, \bm{\beta}, \bm{\nu}, \bm{s} | \bm{Y}, \bm{z}, \bm{\pi})$. To do so, we use the following relation between the gamma function and the Stirling numbers of the first kind denoted by $S$:

\begin{align*}
&\frac{\Gamma(x + n)}{\Gamma(x)} =  (x)_n = \sum_{s=0}^n  S(n,s)(x)^s
\end{align*}

Here $(x)_n$ denotes the Pochhammer symbol. Taking into account also the definition of the beta distribution we reach the following expression:

\begin{equation}
{\begin{split}
&p(\bm{\eta}, \bm{\beta} | \bm{Y}, \bm{z}, \bm{\pi}) \propto  \prod_\ell \prod_t  \left[  \int_0^1 \nu^{\beta_t^{q_\ell} - 1} (1-\nu) ^{n_{\ell t}-1} d\nu \prod_c  \sum_{s=0}^{n_{\ell t c}} S(n_{\ell tc},s) (\beta_t^{q_\ell}\eta_{tc}^{q_\ell})^s  \right] \times \\
&\times \prod_m \prod_t \left[ \frac{\Gamma(\phi_t)}{\prod_c \Gamma(\phi_t \gamma_{tc})} \prod_c \left( \eta_{tc}^m \right) ^{\phi_t \gamma_{tc} - 1} \right]  \prod_m \prod_t \frac{b_t^{a_t}}{\Gamma(a_t)} (\beta_t^m)^{a_t - 1} \exp (-b_t \beta_t^m) \\
\end{split}}
\end{equation}

And therefore we can introduce a set of auxiliary variables $\bm{\nu}$ and $\bm{s}$ such that the joint distribution is given by

 \begin{equation}
 \label{eq:joint_auxiliary}
{\begin{split}
&p(\bm{\eta}, \bm{\beta}, \bm{\nu}, \bm{s} | \bm{Y}, \bm{z}, \bm{\pi}) \propto  \prod_\ell \prod_t  \left[ \nu_{\ell t}^{\beta_t^{q_\ell} - 1} (1-\nu_{\ell t})^{n_{\ell t}-1} \prod_c   S(n_{\ell tc},s_{\ell tc}) (\beta_t^{q_\ell} \eta_{tc}^{q_\ell})^{s_{\ell tc}}  \right] \times \\
&\times \prod_m \prod_t \left[ \frac{\Gamma(\phi_t)}{\prod_c \Gamma(\phi_t \gamma_{tc})} \prod_c \left( \eta_{tc}^m \right) ^{\phi_t \gamma_{tc} - 1} \right]  \prod_m \prod_t \frac{b_t^{a_t}}{\Gamma(a_t)} (\beta_t^m)^{a_t - 1} \exp (-b_t \beta_t^m) \\
\end{split}}
\end{equation}
 
and such that:
\begin{equation}
p(\bm{\eta}, \bm{\beta} | \bm{Y}, \bm{z}, \bm{\pi}) = \int p(\bm{\eta}, \bm{\beta}, \bm{\nu}, \bm{s} | \bm{Y}, \bm{z}, \bm{\pi}) d\bm{\nu} d\bm{s}
\end{equation}

From Equation \ref{eq:joint_auxiliary} it is straightforward to compute the necessary conditional distributions to implement the Gibbs sampler (See Section \ref{hcBBC_inference}).


\vskip 0.2in
\bibliography{sample}

\end{document}